%% file: main.tex
\documentclass{article}

 \usepackage[main, final]{neurips_2026}

\input{_header}

\usepackage[utf8]{inputenc} 
\usepackage[T1]{fontenc}    
\usepackage{hyperref}       
\usepackage{url}            
\usepackage{booktabs}       
\usepackage{amsfonts}       
\usepackage{nicefrac}       
\usepackage{microtype}      

\title{\scalebox{0.93}{LogicTree-RAG: Logic Tree-guided Retrieval-Augmented} \scalebox{0.93}{Generation for Long-form Patent Drafting}}

\author{%
  Jiaqi Zhu \\
  National University of Singapore\\
  \texttt{jiaqi77@nus.edu.sg} \\
  \And
  Naili Xing \\
  National University of Singapore \\
  \texttt{xingnl@comp.nus.edu.sg} \\
  \AND
  Hexiang Pan \\
  National University of Singapore \\
  \texttt{panh@u.nus.edu} \\
  \And
  Haotian Gao \\
  National University of Singapore \\
  \texttt{gaohaotian@comp.nus.edu.sg} \\
  \And
  Jianwei Yin \\
  Zhejiang University \\
  \texttt{zjuyjw@zju.edu.cn} \\
  \And
  Xiaokui Xiao \\
  National University of Singapore \\
  \texttt{xkxiao@nus.edu.sg} \\
  \And
  Beng Chin Ooi \\
  Zhejiang University \\
  \texttt{ooibc@zju.edu.cn} \\
}

\begin{document}

\maketitle

\input{secs/1.abstract}

\input{secs/2.introduction}

\input{secs/3.preliminaries}

\input{secs/4.methodology}

\input{secs/5.experiment}
\input{secs/6.related_work}
\input{secs/7.conclusion}


\small
\bibliographystyle{unsrt} 
\bibliography{references}

\newpage
\appendix
\input{secs/appendix}


\end{document}

%% file: _header.tex
\usepackage{microtype}
\usepackage{graphicx}
\usepackage{subfigure}
\usepackage{booktabs} 
\usepackage{multirow}
\usepackage{makecell}
\usepackage{colortbl} 
\usepackage{tikz}
\usetikzlibrary{tikzmark}

\definecolor{baseblue}{cmyk}{.04,0,0,0}
\usepackage{hyperref}

\usepackage{algorithm}
\usepackage{algorithmic}

\usepackage{wrapfig}

\usepackage{amsmath}
\usepackage{amssymb}
\usepackage{mathtools}
\usepackage{amsthm}

\usepackage[capitalize,noabbrev]{cleveref}

\theoremstyle{plain}
\newtheorem{theorem}{Theorem}[section]

\theoremstyle{definition}
\newtheorem{definition}[theorem]{Definition}

\theoremstyle{remark}

\usepackage[textsize=tiny]{todonotes}

\usepackage{lipsum}

\usepackage{listings}
\definecolor{promptbg}{RGB}{245,245,240}   
\definecolor{promptgreen}{RGB}{0,140,0}    
\definecolor{promptgray}{RGB}{120,120,120} 
\lstdefinestyle{promptstyle}{
  backgroundcolor=\color{promptbg},
  basicstyle=\ttfamily\scriptsize,
  numbers=left,
  numberstyle=\ttfamily\scriptsize\color{promptgray},
  numbersep=10pt,
  frame=single,
  rulecolor=\color{black!10},
  frameround=tttt,
  xleftmargin=0.8em,
  xrightmargin=0.2em,
  aboveskip=0.8em,
  belowskip=0.8em,
  breaklines=true,
  breakatwhitespace=false,
  columns=fullflexible,
  keepspaces=true,
  showstringspaces=false,
  tabsize=2,
  moredelim=[is][\color{promptgreen}\bfseries]{@@@}{@@@}
}

\usepackage{arydshln}

\newcommand{\blue}[1]{{\color{black}#1}}

\newcommand{\term}[1]{{\color{black}#1}}

\newcommand{\oursystem}{LogicTree-RAG}

\usepackage{enumitem}

\newcommand{\ignore}[1]{}

\usepackage[T1]{fontenc}

\newcommand{\newarrow}[1][]{%
  \begin{tikzpicture}[#1]%
    \draw (0,0.7ex) -- (0,0) -- (0.75em,0);
    \draw (0.55em,0.2em) -- (0.75em,0) -- (0.55em,-0.2em);
  \end{tikzpicture}%
}
\usepackage{threeparttable}

\usepackage{caption}

%% file: secs/1.abstract.tex
\begin{abstract}
Long-form technical text generation underpins knowledge-intensive workflows,
yet remains challenging for large language models (LLMs) due to the need for globally consistent logical structuring and faithful technical reasoning beyond local coherence.
Patent drafting is a canonical instance of this challenge, demanding holistic generation of a legally compliant and technically exhaustive document through sustained multi-expert collaboration.
Existing approaches often focus on partial section generation or rely on manually crafted outlines, limiting scalable automation in realistic settings.
In this work, we
propose \oursystem{}, a logic tree-guided retrieval-augmented generation framework that induces a hierarchical logic tree as a global organizational backbone to organize and ground technical disclosures, without relying on expert-defined drafting priors.
Each node in the logic tree represents a technical element and is constructed through evidence-guided recursive generation.
A hybrid traversal mechanism then maps the logic tree into patent sections, enabling controllable and section-balanced generation.
Extensive experiments show that \oursystem{} consistently improves content quality and language conformity over strong LLM-based baselines and achieves longer structured generation with high token efficiency, demonstrating the effectiveness of logic-centric generation for complex technical document drafting.
\end{abstract}

%% file: secs/2.introduction.tex
\section{Introduction}
\label{sec:intro}
Patents are a cornerstone of intellectual property protection, safeguarding technological innovations and enabling their translation into practical applications~\cite{mansfield1986patents,kalanje2006role}.
Drafting a patent, however, is a highly specialized task that requires both a comprehensive understanding of the underlying technical contribution and a detailed knowledge of the formal and legal conventions of patent documents~\cite{goldstein2005patent,waelde2014contemporary}.
In practice, this process often involves close collaboration between \emph{inventors}, who possess the technical expertise but typically lack familiarity with patent law, and \emph{patent attorneys}, who are well-versed in legal conventions but may not fully understand the technical intricacies of the invention.
Consequently, such multi-expert collaboration often necessitates repeated discussion and iterative refinement, making the drafting process labor-intensive and time-consuming.

Recent advances in large language models (LLMs), which have demonstrated remarkable performance across a wide range of natural language processing (NLP) tasks, open up new possibilities for automating and accelerating patent drafting~\cite{shomee2025survey,ding2025automatic,abs-2507-22387}.
Prior work in this area has primarily focused on generating specific sections of patent applications~\cite{christofidellis2022pgt,lee2024instructpatentgpt,wang2024patentformer,bai2024patentgpt,zuo2024patenteval,ren2025large,yoo2025patentscore}, such as title, abstract, or claims, rather than producing a complete document.
While useful for targeted drafting assistance, such partial generation fails to address the core challenge of producing a logically structured patent application in its entirety, where interdependencies across sections are essential.

While a few recent studies have explored full-patent generation, they typically rely on substantial manual pre-design and drafting efforts.
For instance, AutoPatent~\cite{wang2024autopatent} presupposes user-provided draft information, while Pap2Pat~\cite{pap2pat} assumes a detailed outline of each patent section written in advance. 
In their task setups, both methods approximate such user-provided inputs by extracting supervision from reference patents, which distances their designs from real-world patent drafting conditions and ultimately limits automation and scalability.

To address the above-mentioned problems, we formulate patent drafting as a long-text generation task, focusing on producing a complete and logically structured patent application from a pre-publication research paper serving as a proxy for the invention report (IR)~\cite{pap2pat}.
This formulation captures the practical scenario in many research environments, where research papers frequently serve as the initial disclosure for patent drafting.
%
%
Despite its practical relevance, advancing toward fully automated patent drafting entails several non-trivial challenges.
First, patent applications are inherently long-form technical documents, requiring the generation of substantially long and complete texts that remain challenging for LLMs to sustain~\cite{liu2024longgenbench,wu2025longgenbench}.
Second, beyond producing long outputs, effective patent drafting demands globally consistent logical organization and balanced elaboration, a constraint that becomes increasingly difficult to maintain as generation length grows.
Third, robust automation requires \term{minimizing dependence on manually crafted priors}, as such reliance limits scalability and generalization across drafting scenarios.

To tackle these challenges, we propose \oursystem{}, a logic tree-guided retrieval-augmented generation framework for long-form patent drafting.
Unlike prior methods that depend on flat-generation LLMs with manually crafted outlines or templates, \oursystem{} introduces a structured generative paradigm in which \emph{logic precedes content}: it automatically derives a hierarchical logic tree from the input research paper as a global organizational backbone and progressively transforms it into well-formed patent sections in a principled and controllable manner.

As the first step,
\oursystem{} incorporates a \term{semantic chunking mechanism} that structures the input document into semantically coherent units, which are embedded into a \term{vectorized document knowledge base}.
Upon this foundation, we introduce an \term{evidence-guided recursive generation} process for constructing a logic tree that captures the conceptual structure of the underlying research. This process leverages a novel \term{node-aware semantic search} strategy equipped with complementary expansion and refinement procedures, enabling section-level coherence and structural balance.
Each node in the tree corresponds to a technical element, such as a principle, subtask, or subsystem, and is dynamically grounded by evidence retrieved from the knowledge base.
Finally, a \term{hybrid traversal strategy} transforms the completed logic tree into patent sections, yielding a long-form, logically organized patent draft without requiring expert-defined priors.
By serving as a logic-centric intermediate representation, logic tree bridges factual evidence with global generative control, enabling scalable long-form patent drafting.
%
%
%
%

We summarize our main contributions as follows:
\begin{itemize}[itemsep=0mm,leftmargin=4mm]
\vspace{-2mm}

\item We propose \oursystem{}, a logic-centric generative framework for research-to-patent transformation that produces long-form, logically structured patent drafts without reliance on expert-defined drafting priors.

\item We present an evidence-guided recursive generation approach that constructs a logic tree to model the conceptual dependencies and technical contributions with explicit factual grounding in the source document.

\item We introduce a node-aware semantic search with a discriminative relevance scoring that jointly accounts for semantic similarity and structural distinctiveness, improving balanced elaboration and global structural coherence in patent drafting.

\item Extensive experiments under a curated suite of evaluation metrics demonstrate that \oursystem{} improves technical completeness by \underline{11.03}\% and content fidelity by \underline{6.33}\% over strong LLM-based baselines, while enabling longer structured generation with high token efficiency.

\end{itemize}

%% file: secs/3.preliminaries.tex
\section{Preliminaries}
\label{sec:preliminaries}

\begin{figure}[ht]
\begin{center}
\centerline{\includegraphics[width=\linewidth]{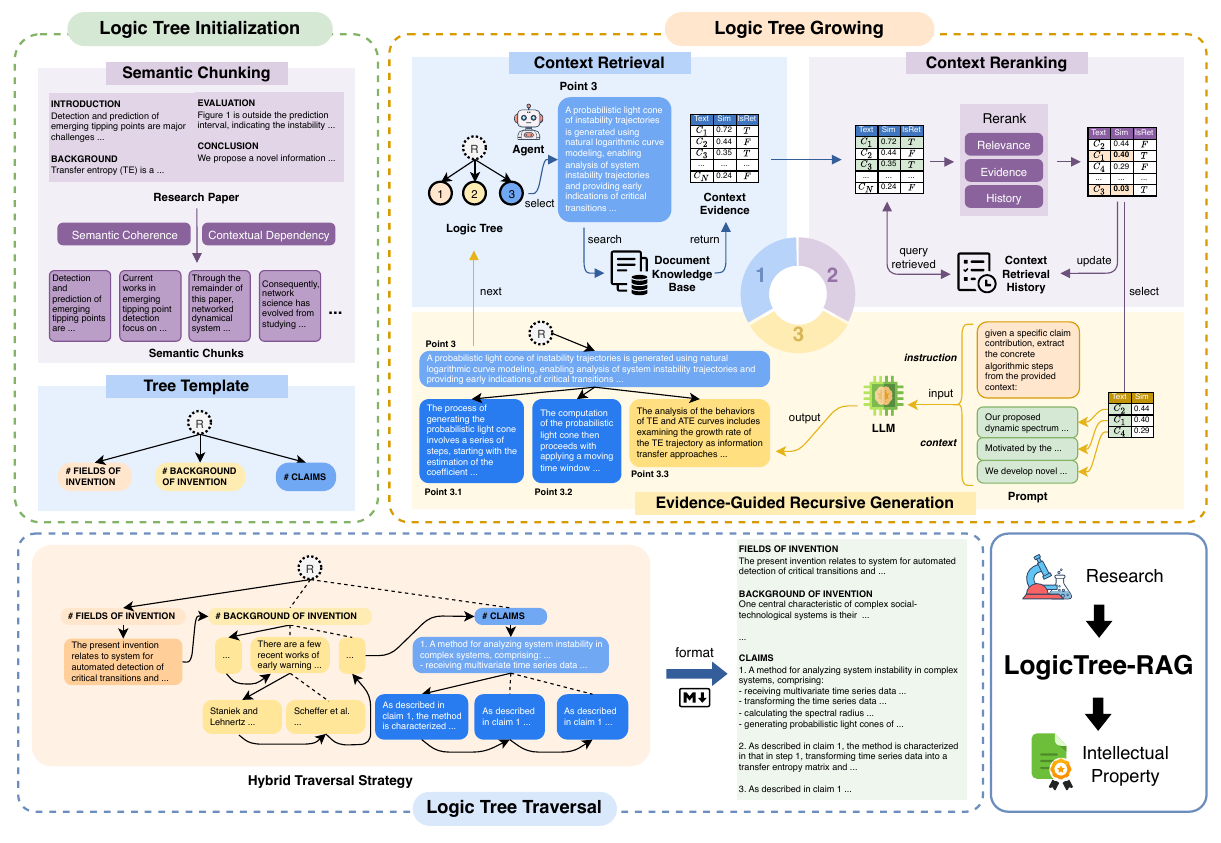}}
\vspace{-1mm}
\caption{Architecture of \oursystem{} framework for long-form patent drafting.}
\label{fig:arch}
\end{center}
\vskip -0.3in
\end{figure}

\textbf{Patents and Patent-Paper Pairs.}
A patent is a structured legal-technical document, typically consisting of \textit{claims}, which define the legally enforceable scope of protection, and a \textit{description}, which discloses the invention in sections such as Background, Summary, and Detailed Description~\cite{jiang2024artificial,pap2pat}.
In many research domains, patents are filed alongside scientific publications, producing \textit{Patent-Paper Pairs}~\cite{murray2002innovation,magerman2010exploring,gans2017contracting,murray2004formal}, where a pre-publication paper serves as the IR for the patent.
Patent-paper pairs align scientific discourse with legal–technical documentation,
offering a valuable yet underexplored resource for studying long-form technical text generation with LLMs.

\textbf{Problem Formulation.}
Consider a patent-paper pair $(X,P)$, where $X = \{x_1, x_2, \dots, x_m\}$ is a pre-publication research paper represented as a collection of text spans, and $P = \{p_1, p_2, \dots, p_n\}$ is the corresponding patent document.
In this work, we formalize the task as generating patent $P$ from paper $X$ mediated by a logic tree $\mathcal{T}$ with evidence grounding, defined as follows.

\begin{definition}[Logic Tree] 
A logic tree is a directed rooted tree $\mathcal{T} = (V,E,r)$, where $V=\{v_1,\cdots,v_N\}$ is the set of nodes, $E\subseteq V \times V$ the set of directed edges specifying parent-child relations, and $r\in V$ the root node representing the overall description of the document.
Each node $v \in V$ is associated with two attributes: (i) a generated text fragment $c(v)$ and (ii) a set of supporting evidence spans $\mathcal{E}(v) \subseteq X$.
\end{definition}
\vspace{-2mm}

Intuitively, the root node summarizes the invention, internal nodes correspond to higher-level sections, and leaf nodes expand into fine-grained technical details. 
To ensure factual grounding, we impose the attribution constraint
\vspace{-1mm}
\begin{align}
\forall v \in V, \quad \mathcal{E}(v) \neq \emptyset,
\end{align}
which requires each generated unit to be anchored to at least one evidence span in the source paper.


%% file: secs/4.methodology.tex
\section{Methodology}
\label{sec:method}
\vspace{-1mm}
\subsection{\oursystem{} Overview}
\vspace{-1mm}
As illustrated in \cref{fig:arch},
\oursystem{} functions as a top-down pipeline that first parses the source document into structured textual units and constructs a \term{logic tree} that captures the conceptual dependencies.
Each node in the tree represents a technical element, such as a principle, step, or subsystem, and is dynamically associated with retrieved evidence from a \term{vectorized document knowledge base}.
\oursystem{} then iteratively expands and refines the tree via \term{evidence-guided recursive generation}, supported by \term{node-aware semantic search} to ensure precise evidence alignment and structural consistency.
Finally, the framework traverses the completed logic tree and hierarchically aggregates node contents into patent sections, yielding a long-form, section-balanced patent draft that integrates technical fidelity with structured generative controllability.

\subsection{Hierarchical Logic Tree Initialization}

To initialize retrievable knowledge units, we utilize layout-aware document analysis~\cite{xu2020layoutlm} to identify hierarchical boundaries across sections and paragraphs, reconstructing the document's logical structure.
Subsequently, we perform semantic chunking to produce contextually coherent segments that from the basis of
the \term{vectorized document knowledge base $\mathcal{V}_{kb}$}.
Unlike fixed chunking methods~\cite{lewis2020retrieval,beltagy2020longformer,huang2023advancing,pap2pat}, our approach dynamically determines chunk boundaries according to the semantic coherence and contextual dependency between adjacent paragraphs.
Formally, given an ordered sequence of paragraphs $X = \{x_1, x_2, \dots, x_m\}$,
we compute the pairwise semantic similarity:
\begin{align}
s_{i,i+1}=\text{sim}(\mathcal{E}_p(x_i),\mathcal{E}_p(x_{i+1})), \; i=1,\cdots,m-1,
\label{eq:pairwise semantic similarity}
\end{align}
where $\mathcal{E}_p(\cdot)$ denotes the paragraph-level embedding function, and $\text{sim}(\cdot,\cdot)$ is instantiated with cosine similarity.
Paragraphs are iteratively merged when $s_{i,i+1} \geqslant \tau_s$ and the accumulated token length does not exceed the upper bound $t_{max}$.
This procedure yields a set of semantically coherent chunks $\tilde{X} = \{\tilde{x}_1, \tilde{x}_2, \dots, \tilde{x}_L\}$
that align with the recovered hierarchy.
Details are provided in Appendix~\ref{sec:appendix-semantic-chunking}.

Following semantic chunking, we initialize a logic tree $\mathcal{T} = (V,E,r)$ to organize and drive the generation process.
To construct an initial structural scaffold, we prompt the backbone LLM with a high-level instruction $\sigma_I$ and the chunk set $\tilde{X}$, generating a draft technical description $\mathcal{D}_d=\mathcal{M}_{LLM}(\sigma_I,\tilde{X})$.
The root node $r$ is instantiated with $c(r)=\mathcal{D}_d$ and $\mathcal{E}(r) = \tilde{X}$, representing the target patent document.
Instead of utilizing syntactic rules or heuristics, we leverage $\mathcal{M}_{LLM}$ to induce the first layer
of the logic tree by producing an evidence-grounded decomposition of key technical components from $\mathcal{D}_d$.
This yields an ordered tree structure in which the root summarizes the invention and its first-level children nodes define the core semantic dimensions to be recursively expanded in subsequent stages.

\subsection{Evidence-Guided Recursive Generation}

Given the initialized logic tree $\mathcal{T}$, we perform recursive generation through a breadth-first expansion strategy.
Specifically, we maintain a FIFO search queue $\mathcal{Q}$ initialized with the leaf nodes of $\mathcal{T}$, ensuring the expansion proceeds in a top–down manner.
For each iteration, we dequeue the front node $v_t$, and retrieve its parent node $v_f$ from $\mathcal{T}$.
The goal of recursive generation is to determine whether $v_t$ should be expanded, refined, or skipped based on the semantic evidence retrieved from the document knowledge base $\mathcal{V}_{kb}$, and to perform the corresponding updates to the logic tree.

To obtain candidate evidence, we perform retrieval over $\mathcal{V}_{kb}$ and collect a set of context spans $\mathcal{D}_{ctx}$, referred to as \term{context evidence}, which are semantically aligned with the node's content $c(v_t)$.
A \term{global novelty test} is then designed to verify whether $\mathcal{D}_{ctx}$ is already covered by other nodes in the logic tree, excluding the child nodes of $v_f$.
If no novel evidence is identified, $v_t$ is skipped;
otherwise, the logic tree is updated via two complementary procedures:
\begin{itemize}[itemsep=0mm,leftmargin=4mm]
\vspace{-2mm}
\item Expansion procedure $E(\cdot)$: Conditioned on $\mathcal{D}_{ctx}$ and an expansion instruction $\sigma_E$, the model generates a supplementary paragraph, which is then segmented into coherent semantic units.
Each unit is instantiated as a child node of $v_t$, forming the set $V_t^{child}=\{v_{tc1},\cdots, v_{tck}\}$, where $c(v_{tci})$ is initialized as the corresponding segmented content and $\mathcal{E}(v_{tci})=\mathcal{D}_{ctx}$.
All nodes in $V_t^{child}$ are then enqueued into $\mathcal{Q}$ for further recursion.


\item Refinement procedure $R(\cdot)$: The model examines $c(v_t)$ to identify under-specified terms using instruction $\sigma_{R1}$.
For domain-specific concepts appearing in $\tilde{X}$, refinement is performed via retrieval-augmented generation with supporting evidence from $\mathcal{V}_{kb}$ and instruction $\sigma_{R2}$; 
otherwise, the model completes the term with instruction $\sigma_{R3}$ based on its internal knowledge.

\vspace{-2mm}
\end{itemize}
Overall, the logic tree $\mathcal{T}$ is updated after each expansion or refinement step. The expansion procedure generates a set of child nodes based on the context evidence, while the refinement procedure resolves under-specified terms and updates the content of the current node, expressed as:
\begin{align}
\label{eq:expansion}
& E(\sigma_E,\mathcal{D}_{ctx})=V_t^{child}\\
& c(v_t) \leftarrow R(\sigma_R,c(v_t),\mathcal{V}_{kb})
\label{eq:refinement}
\end{align}
where $\sigma_R=\{\sigma_{R1},\sigma_{R2},\sigma_{R3}\}$.
The recursive process terminates once either the queue becomes empty or the accumulated content length reaches a predefined budget, ensuring controllable expansion and preventing over-generation.


\subsection{Node-Aware Semantic Search}
\begin{wrapfigure}[16]{t}{0.51\textwidth}
    \begin{center}
    \vspace{-7mm}
\includegraphics[width=0.5\textwidth]{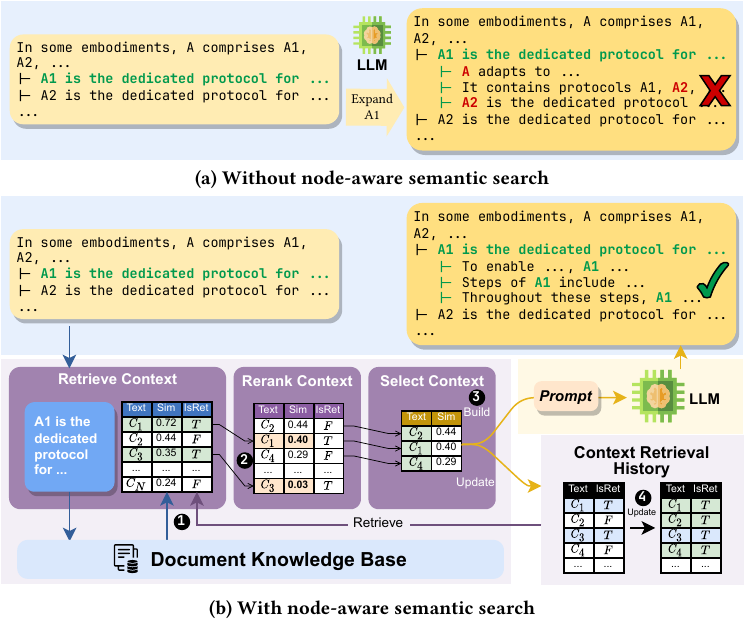}
        \vspace{-2mm}
        \caption{Generation without and with node-aware semantic search.}
        \label{fig:semantic_search}
    \end{center}
\end{wrapfigure}
During the evidence-guided recursive generation, the quality of the context evidence $\mathcal{D}_{ctx}$ critically influences the structural evolution of the logic tree.
Naive retrieval may introduce redundancy or topic drift, particularly when the retrieved spans overlap with already-processed \term{sibling nodes}, leading to incoherent or uneven expansion.
To ensure precision and structural consistency, we impose two desiderata for the context set:
(i) strong semantic association with the current node, and
(ii) low redundancy with its sibling nodes under the same parent.
To meet these requirements, we introduce a node-aware semantic search mechanism.
As illustrated in Figure~\ref{fig:semantic_search}, we first retrieve an initial set of candidate context spans $\tilde{\mathcal{D}}_{ctx}$ from $\mathcal{V}_{kb}$ using the current node content $c(v_t)$ as query:
\begin{align}
\tilde{\mathcal{D}}_{ctx}=\mathcal{R}(c(v_t),\mathcal{V}_{kb})=\text{Top-K}_{a\in \mathcal{V}_{kb}} \text{sim}(a,c(v_t)),
\label{eq:candidate_context}
\end{align}
%
To jointly account for semantic specificity and structural discrimination, we evaluate each candidate span $d \in \tilde{\mathcal{D}}_{ctx}$ through a \term{discriminative relevance score} $S_{dr}(\cdot)$, defined as:
\begin{align}
S_{dr}(d)=\text{sim}(d,c(v_t))-\max_{\scalebox{0.6}{$v_e\in V^{child}_f \setminus \{v_t\}$}} \text{sim}(d,c(v_e)),
\label{eq:filtering_score}
\end{align}
where $V^{child}_f$ denotes the set of child nodes of $v_f$ and $\setminus$ indicates exclusion.
A span $d$ is retained only if $S_{dr}(d)$ exceeds a predefined threshold.
The final context evidence is obtained as $\mathcal{D}_{ctx}=\{d \,|\, S_{dr}(d)\geq \tau_f\}$.
The node-aware semantic search ensures that the retrieved evidence is both node-specific and structurally non-redundant, supporting focused and coherent recursive generation.


\textbf{Remark}.
Although both the node-aware semantic search and the global novelty test operate on evidence relevance and redundancy, they address complementary levels of semantic discrimination.
Node-aware semantic search enforces specificity at the \emph{local sibling level} by filtering evidence overlapping with existing child nodes of $v_t$, while the global novelty test verifies coverage at the \emph{global tree-wide level} by checking whether $\mathcal{D}_{ctx}$ has already appeared elsewhere in the logic tree outside the $v_f$ branch.
Together, these mechanisms provide coordinated local discrimination and global redundancy control, enabling coherent and non-repetitive logic tree construction.

\subsection{\oursystem{} Workflow for Patent Generation}
\label{sec:workflow}

As outlined in Algorithm~\ref{alg:logictree-RAG-gen}, \oursystem{} operates in two stages:
(i) \emph{logic tree construction}, which provides a structured and factually grounded representation of the source document's conceptual space, and
(ii) \emph{patent drafting}, which systematically traverses the constructed logic tree to generate the final patent draft.
The first stage yields a logic tree $\mathcal{T}$ that associates each node $v \in V$ with both generated content $c(v)$ and grounded evidence $\mathcal{E}(v) \subseteq X$, as formalized in \blue{Algorithm~\ref{alg:logictree} in Appendix~\ref{sec:appendix-workflow}}, forming the structural foundation for patent drafting.

In the second stage, \oursystem{} generates the patent draft by traversing the logic tree $\mathcal{T}$ and aggregating node contents.
The hierarchical structure of the logic tree naturally aligns with the organization of complex technical documents, such as patents, where high-level nodes capture overarching concepts, and deeper nodes progressively elaborate technical details, procedural logic, and implementation specifics.
%
To effectively exploit this hierarchical representation, we design a \term{hybrid traversal strategy} that balances content coverage and technical depth across patent sections.
Specifically, high-level sections such as Background and Summary are generated using breadth-first traversal (BFS) to ensure comprehensive and balanced structural coverage.
For the Detailed Description section, BFS is first combined with LLM-based semantic clustering to identify coherent \term{logical subtrees}, each of which is then expanded via depth-first traversal (DFS) to elaborate technical details.
In this way, BFS establishes the structural skeleton, while DFS enables fine-grained expansion, jointly ensuring structural clarity and semantic specificity.
Ultimately, \oursystem{} leverages the logic tree as a structural backbone to automatically transform research documents into long-form patent drafts without relying on manually crafted templates or outlines derived from ground truth.

\begin{algorithm}[t]
\renewcommand{\baselinestretch}{0.9}
\small
\caption{\oursystem{} for Patent Generation}
\label{alg:logictree-RAG-gen} 
\begin{algorithmic}[1]
\renewcommand{\COMMENT}[1]{\footnotesize {\color{gray}\texttt{#1}}}
\renewcommand{\algorithmicrequire}{\textbf{Input:}}  
\renewcommand{\algorithmicensure}{\textbf{Output:}}  
\REQUIRE Pre-publication research paper, the backbone model $\mathcal{M}_{LLM}$,
instruction prompts $\sigma_I, \sigma_E, \sigma_R$.
\ENSURE Generated patent document $P$.

\COMMENT{\scalebox{1}{$\setminus$*Stage I: Logic Tree Construction*$\setminus$}}
\STATE Perform layout-aware semantic chunking on $X$ to obtain chunk set $\tilde{X}$ using Eq.~(\ref{eq:pairwise semantic similarity})
\STATE Initialize Logic Tree $\mathcal{T}$ by generating a draft technical description $\mathcal{D}_d=\mathcal{M}_{LLM}(\sigma_I,\tilde{X})$
\STATE Expand and refine $\mathcal{T}$ via evidence-guided recursive generation with $\sigma_E,\sigma_R$ as Eq.~(\ref{eq:expansion}) and Eq.~(\ref{eq:refinement})

\COMMENT{\scalebox{1}{$\setminus$*Stage II: Patent Drafting*$\setminus$}}
\STATE Initialize empty patent document $P$

\FOR{section $s \in \{\texttt{Background}, \texttt{Summary}\}$}
    \STATE Traverse $\mathcal{T}$ in BFS order over nodes assigned to section $s$
    \STATE Generate section text conditioned on the content $c(v)$ of visited nodes using $\mathcal{M}_{LLM}$
    \STATE Append generated content to $P$
\ENDFOR

\STATE Identify subtrees $\{\mathcal{T}_d\}$ allocated to the \texttt{Detailed Description} section

\FOR{each subtree $\mathcal{T}_d$}
    \STATE Traverse $\mathcal{T}_d$ in DFS order
    \STATE Generate fine-grained technical content using $\mathcal{M}_{LLM}$
    \STATE Append generated content to $P$
\ENDFOR

\STATE {\bfseries Return} Generated patent document $P$

\end{algorithmic}
\end{algorithm}

\ignore{
During traversal, node contents are assigned to their corresponding patent sections and concatenated according to section-specific formatting rules (e.g., section order and length).
Sibling nodes within the same level form parallel subsections, while parent–child relationships translate into hierarchical narrative flow.
Ultimately, \oursystem{} leverages the logic tree as a structural backbone to achieve automatic transformation from research document to long-form, coherent, and balanced patent, without relying on manually crafted templates or expert-defined drafting rules.
}


\blue{\textbf{Remark}}.
\oursystem{} enables scalable and controllable long-form generation through an explicit hierarchical intermediate representation. By decomposing the source document into a logic tree and constraining generation to structurally relevant subtrees, the computational cost is bounded by the tree size and traversal budget, enabling fine-grained control over content expansion. Moreover, by localizing evidence retrieval via node-aware semantic search, \oursystem{} reduces redundant context consumption and improves token efficiency for long and complex technical documents.

%% file: secs/5.experiment.tex
\vspace{-2mm}
\section{Experiments}
\label{sec:exps}
\vspace{-2mm}

\subsection{Experimental Setup}
\label{sec:exps-setup}
\vspace{-1mm}
\noindent \textbf{Benchmarks.}
We first conduct experiments on Pap2Pat~\cite{pap2pat}, a recently proposed benchmark for long-form patent generation based on paper-patent pairs.
Pap2Pat consists of 1813 aligned scientific papers and their corresponding granted patents across various domains.
To further evaluate the generalizability of our framework beyond patent drafting, we additionally consider the LCFO benchmark~\cite{costa2025lcfo}, which focuses on long-context summarization and document expansion.
LCFO consists of 252 long documents ($\sim$5k words each) spanning multiple domains, each paired with multi-level summaries and QA annotations, enabling controlled evaluation of long-form expansion.

\vspace{-1mm}
\noindent \textbf{Baselines.}
We compare \oursystem{} with 13 baselines across three categories:
(i) \emph{Heuristic Skyline}, including ground-truth patents as an oracle upper bound and a case where the source paper is treated as the generated document~\cite{pap2pat}.
(ii) \emph{Single LLM-Call}, where open and proprietary long-context LLMs (Kimi K2~\cite{team2025kimi}, Qwen3-80B, Qwen3-Max~\cite{abs-2505-09388}, DeepseekV3~\cite{abs-2412-19437}, and GPT-5~\cite{achiam2023gpt} and Gemini-2.5 Pro~\cite{comanici2025gemini} under deep thinking mode) are prompted to generate patents from the full paper.
(iii) \emph{Agentic Framework}, including COPGEN~\cite{pap2pat}, along with its SFT-tuned variant, LongWriter~\cite{bai2025longwriter}, and LongWriter-Zero~\cite{wu2026longwriter}, as well as agentic paradigms ReAct~\cite{yao2023react} and Plan-then-Execute~\cite{he2025plan} for step-wise planning and tool use. 
See Appendix~\ref{sec:appendix-baselines} for baselines and implementation details.

\vspace{-1mm}
\noindent \textbf{Implementation.}
\oursystem{} is implemented in Python with Pytorch 2.0.1, with Milvus v2.4.7 as the document knowledge base for document indexing and retrieval.
All methods are evaluated under identical LLM decoding configurations to ensure a fair comparison.
Detailed prompts and implementation settings are provided in Appendices~\ref{sec:appendix-prompt} and \ref{sec:appendix-implementation}.


\vspace{-1mm}
\noindent \textbf{Evaluation Metrics.}
We evaluate patent generation from both content-level and language-level perspectives, complemented by a human expert study conducted by patent practitioners.
\textit{Content-level} metrics assess whether the generated patent conveys accurate and sufficient technical information grounded in the reference materials, including Coverage (NLI-based SCALE~\cite{lattimer2023fast}), Factuality (patent-grounded $\mathcal{F}_{Pat}$ and source-grounded $\mathcal{F}_{Src}$), and Semantic Similarity (ROUGE-L~\cite{lin2004rouge}).
\textit{Language-level} metrics evaluate adherence to patent writing conventions, including Style (StyloMetrix~\cite{okulska2023stylometrix}), Repetition ($\mathcal{R}_r$~\cite{cettolo2014repetition}), and Coherence (DiscoScore~\cite{zhao2023discoscore}).
Details are provided in Appendix~\ref{sec:appendix-evaluation-metrics}.

\begin{table}[t]
   \small
    \centering
    \vspace{-2mm}
    \renewcommand{\arraystretch}{1.1}
    \caption{Overall performance comparison of long-form patent drafting.
    The best results are highlighted in \underline{\textbf{bold}}, and the second-best results are in \textbf{bold}.
    }
    \label{tab:exp-main}
    \resizebox{1\columnwidth}{!}{
        \begin{tabular}{l ccccc ccc c}
\toprule
\tikzmark{tl} \multirow{3}{*}{Method} & \multirow{3}{*}{\makecell[c]{ Output\\ Tokens}} & \multicolumn{4}{c}{Content-level} & \multicolumn{3}{c}{Language-level}  & \multirow{3}{*}{\makecell[c]{Time (s)\\ / Patent (Token)}} \\
 \cmidrule(lr){3-6} \cmidrule(lr){7-9}

& & \multirow{2}{*}{Coverage $\uparrow$} & \multicolumn{2}{c}{Factuality $\uparrow$} & \multirow{2}{*}{\makecell[c]{Semantic \\ $\,\,$ Similarity $\uparrow \,\,$}} & \multirow{2}{*}{Style $\uparrow$} & \multirow{2}{*}{\makecell[c]{Repetition \\ $\mathcal{R}_r$ $\downarrow$}} & \multirow{2}{*}{Coherence $\uparrow$}\\
\cmidrule(lr){4-5} 
& & & $\,\,\,\, \mathcal{F}_{Pat} \,\,\,\,$ & $\,\,\,\, \mathcal{F}_{Src} \,\,\,\,$ &  & &  & \tikzmark{br}\\
  \toprule[0.8pt]
  
\multicolumn{10}{c}{Heuristic Skyline}\\
Reference Patent (GT)  & 18.16k & 89.05 & 88.74 & 88.80 & 100 & 100 & 13.91 & 100 & - \\
Source Paper & 8.03k & 44.77 & 46.06 & 88.68 & 39.37 & 39.26 & 8.39 & 98.60 & -\\
\midrule
\multicolumn{10}{c}{Single LLM-Call Generation} \\
Kimi K2 & 3.26k & 38.21 & 42.76 & 58.95 & 21.28 & 39.47 & 12.04 & 95.92 & 64 (19.63) \\

Qwen3-80B  & 8.91k & 39.09 & 37.50 & 44.88 & 34.08 & 35.06 & 6.05 & 95.71 & 81 (9.10) \\

Qwen3-Max & 8.91k & 39.44 & 37.68 & 45.22 & 34.09 & 35.06 & 8.41 & 97.14 & 172 (19.29)\\

DeepseekV3 & 1.87k & 36.70 & 47.09 & 59.66 & 11.98 & 32.68 & 6.13 & 95.56 & 43 (22.89)\\
\hdashline
GPT-5 DT & 2.95k & 40.91 & 46.70 & 62.04 & 20.56 & 41.52 & 7.25 & 97.34 & 146 (49.49)  \\

Gemini-2.5 Pro DT & 2.69k & 38.69 & 47.58 & 60.97 & 24.25 & 39.68 & 6.09 & 97.12 & 109 (40.52) \\

\midrule
\multicolumn{10}{c}{Agentic Framework Generation} \\

COPGEN & 7.69k & 39.21 & 57.63 & 62.39 & 32.56 & 59.54 & 15.47 & 97.11 & 265 (34.46)\\
COPGEN$^\dagger$ & 8.96k & 32.85 & 52.44 & 58.16 & 28.96 & 43.38 & 18.94 & 97.21 & 216 (24.11)\\
\newarrow w/ SFT & 25.85k & 40.77 & 50.12 & 59.69 & 38.72 & 62.21 & 21.58 & \underline{\textbf{97.50}} & 221 (8.55)\\

\hdashline
LongWriter & 10.81k & 37.21 & 36.65 & 45.21 & 21.12 & 34.26 & 6.67 & 95.62 & 197 (18.22) \\
LongWriter-Zero & 12.16k & 36.92 & 38.36 & 46.17 & 22.86 & 36.72 & 5.76 & 96.84 & 258 (21.11) \\
\hdashline

\rowcolor{blue!6}\oursystem{} & 20.22k & \underline{\textbf{43.79}} & \underline{\textbf{58.92}}  & \textbf{66.34} & \textbf{38.91} & \underline{\textbf{65.68}} & \underline{\textbf{4.14}} & \textbf{97.45} & 311 (15.38) \\
\rowcolor{blue!6} \newarrow Contamination Control  &  18.96k & \textbf{42.95} & \textbf{57.79}  & \underline{\textbf{66.57}} & \underline{\textbf{40.28}} & \textbf{64.28} & \textbf{4.60} & 97.23 & 298 (15.72) \\
\bottomrule

        \end{tabular}
        }
    \vspace{-3mm}
\end{table}

\begin{figure}[t]
  \centering
  \begin{minipage}[t]{0.47\textwidth}
    \centering
\includegraphics[width=\textwidth]{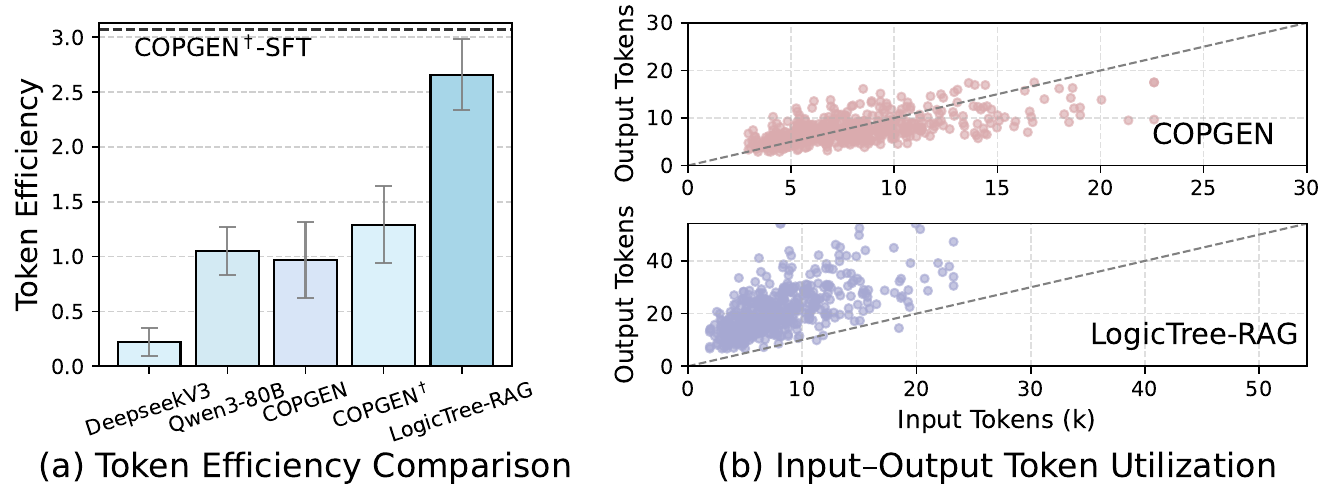}
\vspace{-6mm}
    \caption{\small Token efficiency comparison results.}
\label{fig:exp-tokens}
  \end{minipage}
  \hfill
  \begin{minipage}[t]{0.52\textwidth}
    \centering
\includegraphics[width=\textwidth]{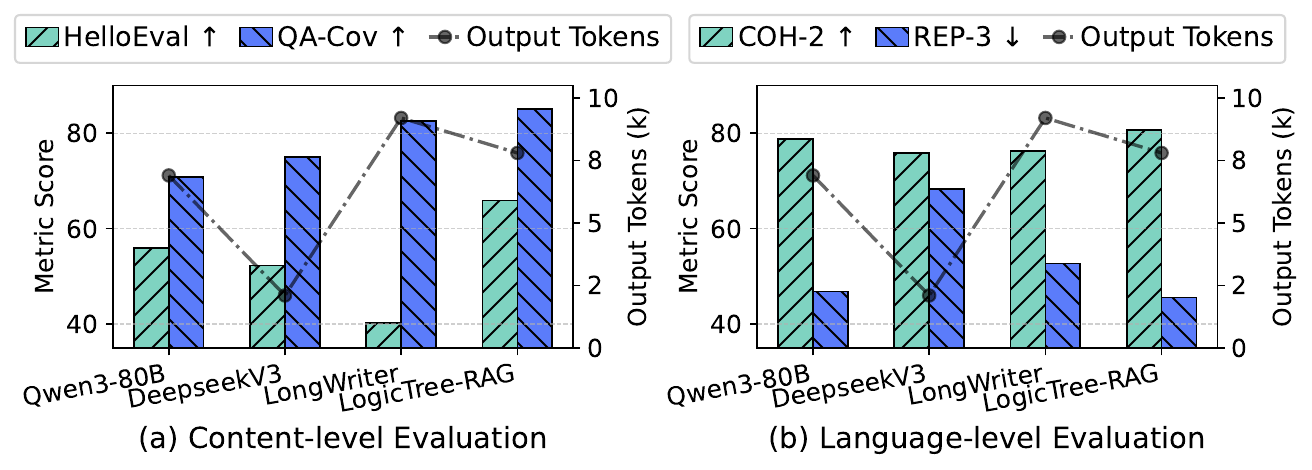}
\vspace{-6mm}
    \caption{\small Generalization performance on LCFO.}
    \label{fig:generalizability}
  \end{minipage}
  \vspace{-5mm}
\end{figure}

\vspace{-1mm}
\subsection{Main Results}
\label{sec:exps-main}
\vspace{-1mm}

The overall results are summarized in Table~\ref{tab:exp-main}. \oursystem{} delivers consistently superior performance across key content and language metrics, generating substantially longer patent drafts that outperform strong long-context baselines.
%
At the \textit{content level}, \oursystem{} achieves the highest coverage, factuality, and semantic similarity among all baselines.
These results underscore that logic-tree-guided generation improves technical completeness and evidence-grounded elaboration beyond single-call long-context prompting and agentic frameworks even with post-training.
%
At the \textit{language level}, \oursystem{} demonstrates strong performance in stylistic conformity and repetition rate. Its coherence score ranks second only to the SFT-tuned COPGEN variant, while requiring no task-specific fine-tuning.
%
To control for potential training-data contamination, we evaluate \oursystem{} on a temporally filtered post-2024 subset beyond typical pretraining cutoffs, observing consistent gains without degradation, suggesting that the improvements are not driven by training-data overlap.
Moreover, \oursystem{} maintains competitive per-token efficiency with only modest overhead, offering a favorable quality–efficiency trade-off for long-form patent drafting.

\ignore{
Moreover, we report the average generation time per patent and per token.
Despite its structured workflow, \oursystem{} incurs only modest overhead while maintaining competitive per-token efficiency,
offering a favorable quality–efficiency trade-off.
}

\begin{figure}[t]
\begin{center}
\centerline{\includegraphics[width=\linewidth]{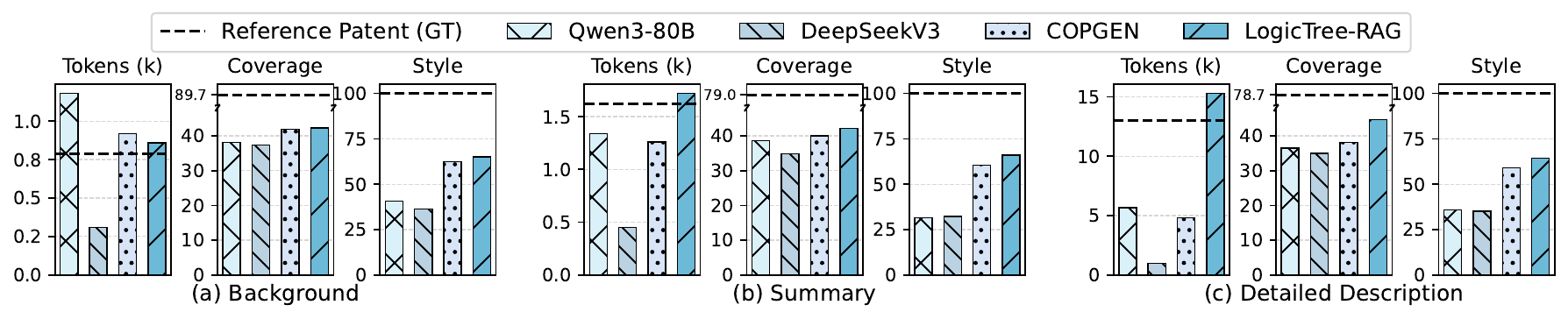}}
\vskip -0.1in
\caption{Section-level performance
in terms of output length, content coverage, and linguistic style.}
\label{fig:exp-section-level}
\end{center}
\vskip -0.2in
\end{figure}

\ignore{
\begin{figure}[t]
\begin{center}
\centerline{\includegraphics[width=0.5\linewidth]{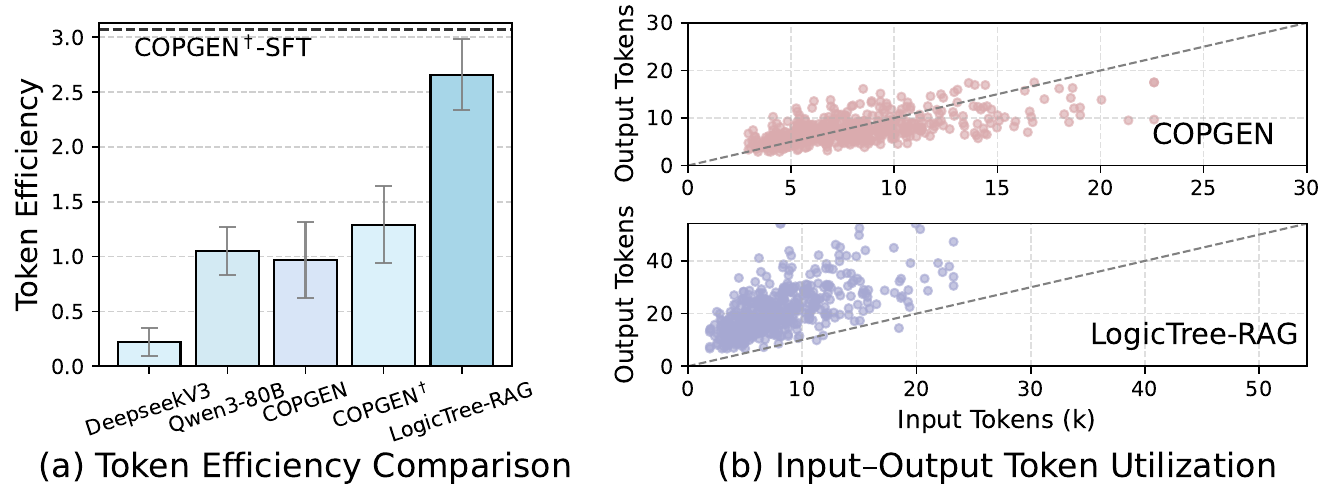}}
\vspace{-1mm}
\caption{
Token efficiency comparison for long-form patent drafting.}
\label{fig:exp-tokens}
\end{center}
\vskip -0.4in
\end{figure}
}

\vspace{-1mm}
\subsection{In-Depth Analysis of Patent Generation}
\vspace{-1mm}

\textbf{Token Efficiency.}
We further analyze token efficiency, defined as the ratio of generated output tokens to total input tokens consumed across all LLM invocations, which serves as a proxy for both generation efficiency and scalability.
As illustrated in Figure~\ref{fig:exp-tokens}(a), \oursystem{} achieves the highest token efficiency among all baselines, closely approaching the SFT-enhanced COPGEN variant despite requiring no task-specific fine-tuning.
Figure~\ref{fig:exp-tokens}(b) further shows that \oursystem{} shifts the input–output token distribution toward a more efficient regime and effectively converts contextual input into useful long-form patent content.
Detailed cost breakdowns are provided in Appendix~\ref{sec:appendix-cost}.

\textbf{Generalizability to Long-Text Generation.}
\oursystem{} is not tied to patent drafting, but targets a broader setting of source-grounded structured long-form generation, where the goal is to transform a long source document into a coherent and factually grounded target document.
To validate this generalizability, we evaluate \oursystem{} on the LCFO benchmark~\cite{costa2025lcfo} for document expansion.
Following~\cite{costa2025lcfo}, we report output tokens, coherence (COH-2), repetition (REP-3), overall quality via an LLM-as-a-judge metric (HelloEval), and human-evaluated QA-based coverage (QA-Cov), which measures the fraction of benchmark questions answerable from the generated output.
Without modifying the core pipeline, we apply \oursystem{} by adapting only the output schema. 
As shown in Figure~\ref{fig:generalizability}, \oursystem{} consistently outperforms baselines in all content and language-level metrics, while maintaining competitive output length.
These findings suggest that the benefits of logic-tree-guided generation extend beyond patent drafting to general long-form generation tasks.

\vspace{-1mm}
\textbf{Section-Level Analysis.}
We analyze generation quality across patent sections, i.e., Background, Summary, and Detailed Description, considering length, content coverage, and linguistic style, with results summarized in Figure~\ref{fig:exp-section-level}.
\oursystem{} produces section-wise output lengths closely aligned with ground-truth patents, indicating effective control over generation granularity and mitigating the length imbalance observed in single-call and chunk-based baselines.
From a content perspective, \oursystem{} achieves the highest coverage across all patent sections, with up to a 17.4\% improvement in the Detailed Description, where fine-grained technical completeness is critical and competing methods tend to under-generate.
For language quality, \oursystem{} consistently attains the best style scores, demonstrating strong stylistic consistency across sections.

\begin{figure}[t]
  \centering
  \begin{minipage}[t]{0.5\textwidth}
    \centering
\includegraphics[width=1\textwidth]{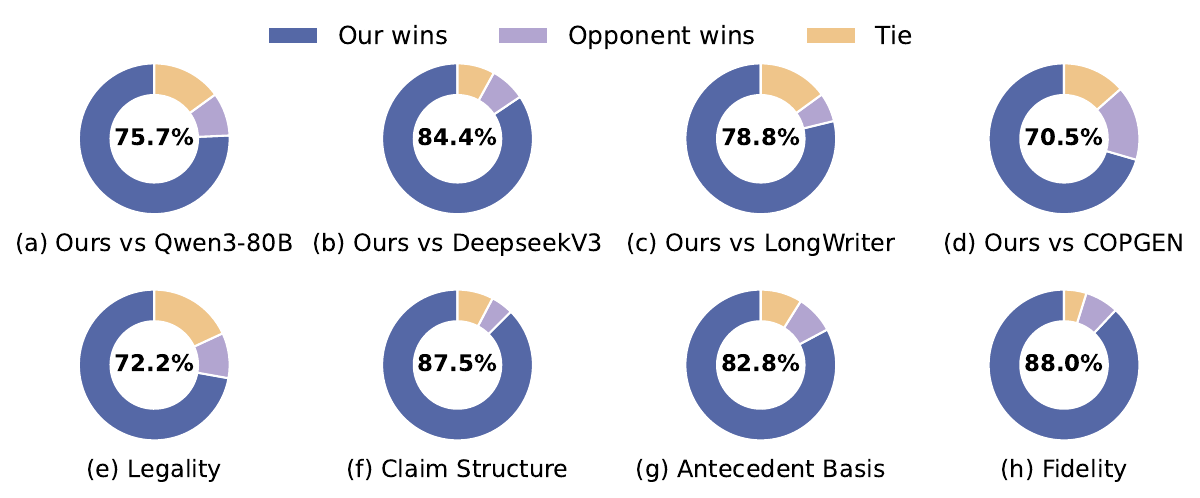}
\vspace{-6mm}
    \caption{\scalebox{0.91}{\small Win-rate results (\%) under pairwise comparison.}
    }
\label{fig:exp-expert}
  \end{minipage}
  \hfill
  \begin{minipage}[t]{0.49\textwidth}
    \small
    \centering
    \renewcommand{\arraystretch}{1.08}
    \vspace{-30mm}
    \captionof{table}{\scalebox{0.95}{\small Component-wise ablation of \oursystem{}.}}
    \vspace{-1mm}
    \label{tab:component-wise-ablation}
\resizebox{\columnwidth}{!}{
        \begin{tabular}{l c c c c c}
    \toprule[1.5pt]
\multirow{2}{*}{Variant} & \multirow{2}{*}{\makecell[c]{Output \\Tokens}} & \multicolumn{2}{c}{Content-level} & \multicolumn{2}{c}{Language-level} \\
\cmidrule(lr){3-4} \cmidrule(lr){5-6}
 & & Coverage & Factuality ($\mathcal{F}_{Pat}$) & Style & RR ($\mathcal{R}_r$)\\
    \midrule[0.5pt]
  w/o Logic Tree & 9.10k & 39.85 & 56.92 & 60.47 & 16.86 \\
Logic Tree w/o Retrieval & 8.21k & 40.56 & 51.16 & 59.05 & 9.38 \\
     \midrule[0.1pt]
   w/o Semantic Chunking & 19.42k & 42.09 & 53.74 & 62.27 & 10.93 \\
   \hdashline
   w/o Refinement & 18.58k & 39.92 & 56.85 & 61.87 & \textbf{4.09} \\
   w/o Evidence-Grounded $\sigma_{R2}$& 19.60k & 40.16 & 56.98 & 62.64 & 4.67 \\
   w/o Internal Knowledge $\sigma_{R3}$ & 18.95k & 43.58 & 58.85 & 65.62 & 5.21  \\
   \hdashline
    w/o Node-aware Reranking & 20.85k & 42.60 & 57.13 & 64.79 & 8.32 \\
    w/ DFS Traversal & 21.13k & 41.15 & 58.68 & 65.12 & 5.34 \\
    \midrule[0.1pt]
   \oursystem{} & 20.22k & \textbf{43.79} & \textbf{58.92} & \textbf{65.68} & 4.14 \\
    \bottomrule[1.5pt]
        \end{tabular}
    }
  \end{minipage}
  \vspace{-4mm}
\end{figure}

\vspace{-1mm}
\textbf{Human Expert Review.}
To assess practical applicability beyond automatic metrics, we conduct a blinded human expert evaluation with three patent practitioners using five criteria (i.e., Overall Quality, Legality, Claim Structure, Antecedent Basis, and Fidelity).
Experts perform pairwise comparisons between our method and baselines.
Figure~\ref{fig:exp-expert}(a-d) present pairwise comparisons against individual baselines, while (e-h) summarize performance across evaluation criteria. 
The results show that \oursystem{} consistently achieves strong human preference across all dimensions, indicating its practical utility in real-world scenarios.
Full experimental details are provided in Appendix~\ref{sec:appendix-expert}.

\ignore{
\begin{table}[t]
   \small
    \centering
    \renewcommand{\arraystretch}{1.2}
    \caption{Component-wise ablation of \oursystem{}.}
    \vspace{-2mm}
    \label{tab:component-wise-ablation}
\resizebox{0.5\columnwidth}{!}{
        \begin{tabular}{l c c c c c}
    \toprule[1.5pt]
\multirow{2}{*}{Variant} & \multirow{2}{*}{\makecell[c]{Output \\Tokens}} & \multicolumn{2}{c}{Content-level} & \multicolumn{2}{c}{Language-level} \\
\cmidrule(lr){3-4} \cmidrule(lr){5-6}
 & & Coverage & Factuality ($\mathcal{F}_{Pat}$) & Style & RR ($\mathcal{R}_r$)\\
    \midrule[0.5pt]
  w/o Logic Tree & 9.10k & 39.85 & 56.92 & 60.47 & 16.86 \\
Logic Tree w/o Retrieval & 8.21k & 40.56 & 51.16 & 59.05 & 9.38 \\
     \midrule[0.1pt]
   w/o Semantic Chunking & 19.42k & 42.09 & 53.74 & 62.27 & 10.93 \\
    w/o Node-aware Reranking & 20.85k & 42.60 & 57.13 & 64.79 & 8.32 \\
    w/ DFS Traversal & 21.13k & 41.15 & 58.68 & 65.12 & 5.34 \\
    \midrule[0.1pt]
   \oursystem{} & 20.22k & \textbf{43.79} & \textbf{58.92} & \textbf{65.68} & \textbf{4.14} \\
    \bottomrule[1.5pt]
        \end{tabular}
    }
     \vspace{-5mm}
\end{table}
}

\vspace{-1mm}
\subsection{Ablation and Sensitivity Analysis}


\vspace{-1mm}
\textbf{Component-wise Ablation.}
We evaluate ablated variants of \oursystem{} with all other settings fixed, as shown in Table~\ref{tab:component-wise-ablation}.
Removing the logic tree or retrieval substantially shortens the outputs and degrades content quality, causing a 9.0\% drop in coverage and a 13.2\% degradation in factuality, respectively. This confirms the necessity of structural organization and evidence grounding.
Removing semantic chunking or node-aware reranking weakens factuality and increases repetition, highlighting the importance of coherent and non-redundant evidence selection.
Moreover, removing evidence-grounded refinement $\sigma_{R2}$ leads to consistent degradation across all metrics, whereas removing internal-knowledge refinement $\sigma_{R3}$ yields negligible impact.
Since $\sigma_{R3}$ is triggered in only 8.6\% of steps, it mainly serves as a fallback for under-specified cases, while the main gains come from retrieved evidence.
Replacing hybrid traversal with pure DFS produces longer but less comprehensive outputs, indicating that the hybrid strategy better balances breadth and depth.
Overall, the full \oursystem{} achieves the best performance across metrics, validating the joint contribution of structural modeling, evidence grounding, and traversal design.

\begin{wrapfigure}[7]{t}{0.5\textwidth}
    \begin{center}
    \vspace{-7mm}
\includegraphics[width=0.48\textwidth, height=2.3cm]{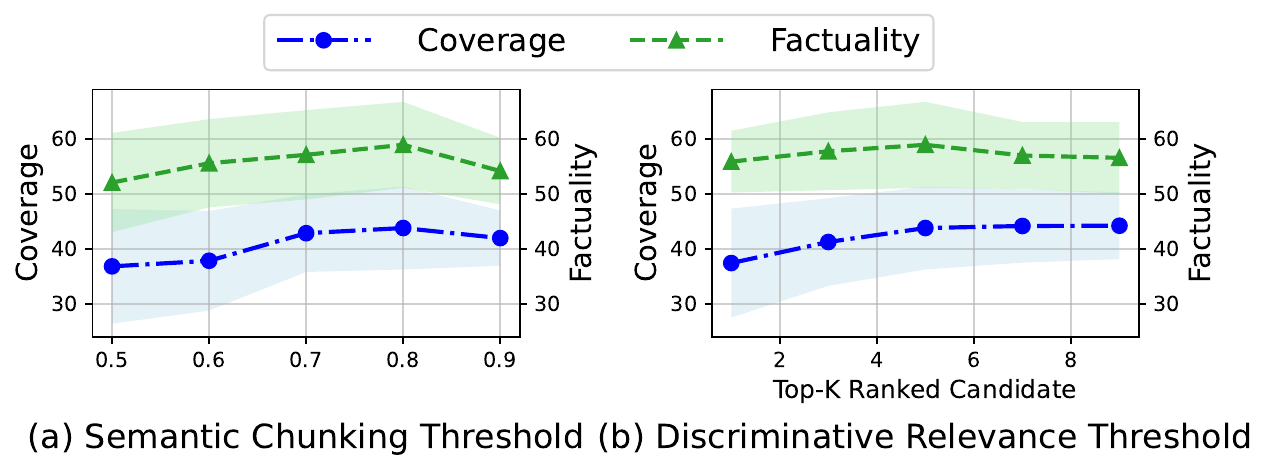}
        \vspace{-2mm}
        \caption{Hyperparameter Sensitivity Analysis.}
        \label{fig:exp-sensitivity}
    \end{center}
\end{wrapfigure}
\vspace{-1mm}
\textbf{Sensitivity to Key Hyperparameters.}
We study the effects of the semantic chunking threshold $\tau_s$ and the discriminative relevance threshold controlled by the ranked candidate $k$.
As shown in Figure~\ref{fig:exp-sensitivity}(a), both coverage and factuality improve as $\tau_s$ increases from low to moderate values, and peak around $\,$0.7--0.8, $\,$suggesting that overly coarse chunks introduce noisy evidence, $\,$while overly fine-grained chunks reduce contextual sufficiency.
Figure~\ref{fig:exp-sensitivity}(b) shows that increasing $k$ initially improves coverage by allowing more supporting evidence, but factuality peaks at a moderate value and slightly declines when weaker or redundant evidence is included.
This highlights the importance of balancing evidence sufficiency and discriminative filtering, and supports the default setting of $k=5$ as an effective compromise.
Further analysis of tree growth budget parameters is provided in Appendix~\ref{sec:appendix-treesize}.

We provide additional analyses on backbone choices, agentic pipeline variants, and evidence support in Appendices~\ref{sec:appendix-ablation}, \ref{sec:appendix-agentic}, and~\ref{sec:appendix-evidence}, further supporting the generality and reliability of LogicTree-RAG.


\vspace{-2mm}
\subsection{Broader Implications and Limitations}
\label{sec:exps-limit}
\vspace{-2mm}

This work advances controllable and evidence-grounded generative AI for long-form technical document drafting.
By framing patent drafting as a representative high-stakes structured generation task, LogicTree-RAG can assist practitioners in organizing complex technical disclosures into coherent and structured drafts, potentially reducing manual effort in early-stage drafting.
However, the framework is intended to assist rather than replace domain experts, and its outputs should be treated as preliminary drafts and reviewed by qualified professionals before any formal use.
Beyond patents, the logic-centric paradigm extends naturally to broader long-text generation tasks that require global organization and factual grounding.
While LCFO results support such generalization, further validation on diverse long-form generation settings remains an important direction for future work.

\ignore{
While this work focuses on controllable and evidence-grounded long-form patent drafting, the proposed framework is intended as a structured drafting assistant rather than a replacement for professional patent attorneys.
Critical aspects of real-world patent practice, such as legal adequacy, claim dependency correctness, and antecedent basis consistency, ultimately require expert human judgment.
Moreover, while LCFO results support generalization beyond patent drafting, further validation on other structured long-form tasks, such as standards drafting and software documentation, remains an important direction for future work.
}

%% file: secs/6.related_work.tex
\vspace{-2mm}
\section{Related Work}
\label{sec:related_work}

\vspace{-2mm}
\textbf{Patent-related Benchmarks.} 
Recent progress in LLMs has spurred benchmarking of patent-centric capabilities~\cite{jiang2025natural}.
For example, IPEval~\cite{ipeval} evaluates patent law QA, PatentEval~\cite{zuo2024patenteval} targets claims-to-abstract and next-claim generation, and HUPD-DCG~\cite{hupd-dcg} measures claim generation from patent descriptions.
Complementary to task-specific benchmarks, Pap2Pat~\cite{pap2pat} extends evaluation to long-text patent drafting, reflecting growing interest in document-level generation. 
These benchmarks illustrate the evolution of patent-related evaluation from localized capabilities toward holistic long-form technical writing.

\vspace{-1mm}
\textbf{Patent-related Generation.}
Existing patent generation methods can be broadly grouped into three paradigms.
(i) \emph{Patent-to-patent} generates specific sections conditioned on other parts within the same patent.
For instance, PGT~\cite{christofidellis2022pgt} is pre-trained to support title-to-abstract and abstract-to-claim generation.
PatentLMM~\cite{patentlmm} presents a large multimodal model to write descriptions of patent figures.
\cite{ren2025large} proposes a knowledge-tuning approach for initial concept discovery and content generation.
(ii) \emph{Draft-to-patent} expands patent drafts into patent-form text.
PatentGPT~\cite{bai2024patentgpt} pre-trains a series of base LLMs and benchmarks them with patent generation from technical disclosure documents draft.
AutoPatent~\cite{wang2024autopatent} designs a multi-agent workflow to migrate a form of answers to technical questions to the corresponding patent draft.
(iii) \emph{Paper-to-patent} targets long-form patent drafting from research papers, exemplified by Pap2Pat~\cite{pap2pat} with outline-guided generation.
However, prior methods either operate under partial-context settings or depend on externally provided (often manually crafted) drafting priors, leaving realistic long-form patent drafting with global logical organization less well addressed.
See Appendix~\ref{sec:appendix-related-work} for a comprehensive review.




%% file: secs/7.conclusion.tex
\vspace{-1mm}
\section{Conclusions}
\label{sec:conclusion}
\vspace{-1mm}
Automated patent drafting exposes a fundamental bottleneck in long-form technical text generation, where sustaining globally consistent logical structure and faithful technical reasoning remains challenging.
In this paper, we propose \oursystem{}, a logic tree-guided retrieval-augmented generation framework for long-form patent drafting that introduces a hierarchical logic tree as a global organizational backbone, where each node represents a technical element and is grounded by retrieved evidence via an evidence-guided recursive generation mechanism.
To ensure precise and balanced tree construction, we develop node-aware semantic search with discriminative relevance scoring and design a hybrid traversal strategy that maps the logic tree into structurally coherent patent sections.
Extensive experiments show that \oursystem{} consistently improves content quality and language conformity over strong long-context and multi-stage LLM-based baselines, enabling longer structured generation with high token efficiency.

%% file: secs/appendix.tex
\section*{Appendix}

















\begin{itemize}
  \item \textbf{\ref{sec:appendix-notation}. Notation Table} \dotfill \pageref{sec:appendix-notation}

  \item \textbf{\ref{sec:appendix-related-work}. Extended Related Work} \dotfill \pageref{sec:appendix-related-work}

  \item \textbf{\ref{sec:appendix-semantic-chunking}. Details on Semantic Chunking} \dotfill \pageref{sec:appendix-semantic-chunking}

  \item \textbf{\ref{sec:appendix-workflow}. Workflow of Logic Tree Construction} \dotfill \pageref{sec:appendix-workflow}

  \item \textbf{\ref{sec:appendix-prompt}. Patent Generation Prompt} \dotfill \pageref{sec:appendix-prompt}

  \item \textbf{\ref{sec:appendix-implementation}. Implementation Details} \dotfill \pageref{sec:appendix-implementation}
  \begin{itemize}
      \item[$\cdot$] \textit{\ref{sec:appendix-evaluation-metrics}. Evaluation Metrics} \dotfill \pageref{sec:appendix-evaluation-metrics}
      \item[$\cdot$] \textit{\ref{sec:appendix-baselines}. Baseline Methods} \dotfill \pageref{sec:appendix-baselines}
      \item[$\cdot$] \textit{\ref{sec:appendix-Hyperparameters}. Hyperparameters and Decoding Configurations} \dotfill \pageref{sec:appendix-Hyperparameters}
      \item[$\cdot$] \textit{\ref{sec:appendix-Environment}. Experimental Environment} \dotfill \pageref{sec:appendix-Environment}
      \item[$\cdot$] \textit{\ref{sec:appendix-prompt-settings}. Detailed Prompt Settings} \dotfill \pageref{sec:appendix-prompt-settings}
  \end{itemize}
  
  \item \textbf{\ref{sec:appendix-exps}. Supplementary Experimental Results} \dotfill \pageref{sec:appendix-exps}
  \begin{itemize}
      \item[$\cdot$] \textit{\ref{sec:appendix-ablation}. Supplementary Ablation Study} \dotfill \pageref{sec:appendix-ablation}
      \item[$\cdot$] \textit{\ref{sec:appendix-agentic}. Supplementary Study on Agentic Planning} \dotfill \pageref{sec:appendix-agentic}
      \item[$\cdot$] \textit{\ref{sec:appendix-evidence}. Evidence-support Analysis} \dotfill \pageref{sec:appendix-evidence}
      \item[$\cdot$] \textit{\ref{sec:appendix-cost}. Computational Cost Analysis} \dotfill \pageref{sec:appendix-cost}
      \item[$\cdot$] \textit{\ref{sec:appendix-expert}. Human Expert Study} \dotfill \pageref{sec:appendix-expert}
      \item[$\cdot$] \textit{\ref{sec:appendix-treesize}. Sensitivity to Tree Growth Budget} \dotfill \pageref{sec:appendix-treesize}
      \item[$\cdot$] \textit{\ref{sec:appendix-case}. Case Study} \dotfill \pageref{sec:appendix-case}
  \end{itemize}
\item \textbf{\ref{sec:appendix-discussion}. Discussion on Human Evaluation and Legal Adequacy} \dotfill \pageref{sec:appendix-discussion}
\end{itemize}


\newpage
\section{Notation Table}
\label{sec:appendix-notation}
To provide a comprehensive overview of the notations used throughout the paper, we present a summary of notations in Table~\ref{tab:notation} as a quick reference to facilitate the understanding and recall of each symbol.


\begin{table}[ht]
   \small
    \centering
    \renewcommand{\arraystretch}{1.3}
    \caption{Notations.}
    \label{tab:notation}
    \resizebox{1\columnwidth}{!}{
        \begin{tabular}{c l}
    \toprule[1.5pt]
 Notation & Description\\
  \toprule[0.8pt]
$(X,P)$ &  A patent-paper pair, where $X$ is the source (pre-publication) research paper and $P$ is the corresponding patent document. \\
$X = \{x_1, x_2, \dots, x_m\}$ & Source paper represented as a collection of $m$ text spans. \\
$P=\{p_1,\dots,p_n\}$ & Patent document represented as a collection of $n$ text spans. \\
$\tilde{X} = \{\tilde{x}_1, \tilde{x}_2, \dots, \tilde{x}_L\}$ &  Semantic chunks produced from $X$ after layout-aware semantic chunking, with $L \le m$.\\
$\mathcal{T} = (V,E,r)$ &  Logic tree, a directed rooted tree with node set $V$, edge set $E \subseteq V\times V$, and root node $r$.\\
$V=\{v_1,\cdots,v_N\}$ & Set of logic tree nodes, each representing a technical element (e.g., principle, step, subsystem). \\
$v_t$ & Current node dequeued for recursive generation at iteration $t$. \\
$v_f$ & Parent node of $v_t$ in the logic tree. \\
$V_f^{\text{child}}$ & Set of child nodes of $v_f$. \\
$c(v)$ & Generated text fragment associated with node $v$. \\
$\mathcal{E}(v)$ & Supporting evidence spans from the source paper attributed to node $v$, $\mathcal{E}(v)\subseteq X$. \\
$\mathcal{V}_{kb}$ & Vectorized document knowledge base built from semantic chunks for retrieval. \\
$\mathcal{E}_p(\cdot)$ & Paragraph-level embedding function used for semantic similarity and indexing. \\
$\mathcal{M}_{LLM}$ & Backbone large language model. \\
$\sigma_I$ & Instruction prompt for generating an initial draft technical description. \\
$\mathcal{D}_d$& Draft technical description generated by $\mathcal{M}_{LLM}(\sigma_I,\tilde{X})$ to initialize the logic tree root.  \\
$E(\cdot)$ & Expansion procedure. \\
$R(\cdot)$ & Refinement procedure. \\
$\sigma_E$ & Expansion instruction prompt used to generate new child nodes from retrieved context. \\
$\sigma_R$ & Refinement instruction set; in the draft $\sigma_R=\{\sigma_{R1},\sigma_{R2},\sigma_{R3}\}$. \\
$\mathcal{Q}$ & FIFO queue used for breadth-first expansion over leaf nodes in recursive generation. \\
$\mathcal{D}_{ctx}$ & Context evidence set used to expand or refine the current node $v_t$. \\
$S_{dr}(d)$ & Discriminative relevance score for candidate span $d$. \\
$\tau_f$ & Threshold for retaining a candidate span via discriminative relevance. \\
$\tau_s$ & Semantic chunking threshold. \\
$t_{\max}$ & Upper bound on accumulated token length for a chunk during semantic chunking. \\

$T_d$ & A subtree allocated to the Detailed Description section. \\

    \bottomrule[1.5pt]
        \end{tabular}
        }
\end{table}


\section{Extended Related Work}
\label{sec:appendix-related-work}

In this section, we provide an extended review of related work, encompassing both general long-text generation methods and patent-specific drafting paradigms.

\subsection{Long-text Generation}
Recent advances in long-text generation have explored structured prompting, hierarchical decomposition, and personalization to improve coherence over extended outputs~\cite{wu2026longwriter}.
Early approaches, such as recursive and outline-based generation~\cite{yang2022re3,yang2023doc}, introduce hierarchical planning to maintain narrative consistency across long contexts.
More recent work further incorporates personalization~\cite{kumar2024longlamp,xu2025personalized} and reasoning-oriented training~\cite{salemi2025reasoning}, enabling models to better align outputs with user intent.

With growing demand for ultra-long generation (e.g., beyond 2k tokens), new datasets and training paradigms have emerged.
For instance, large-scale instruction-following datasets constructed via back-translation support long-context learning but often rely on synthetic supervision and remain limited in output length~\cite{pham2025clipper,pham2406suri}. 
Other approaches, such as LongWriter~\cite{bai2025longwriter}, leverage supervised and preference-based optimization on agent-generated long-form data to extend output length, but may inherit biases from teacher models and lack explicit control over global structure.
LongWriter-Zero~\cite{wu2026longwriter} further explores reinforcement learning from scratch to induce ultra-long generation without relying on synthetic supervision, improving length control and structural consistency.
Overall, despite notable progress, existing approaches lack explicit global structure modeling and fine-grained evidence grounding, limiting their ability to ensure coherent, factual, and controllable generation over ultra-long contexts.


Beyond long-form generation, domain-specific LLM systems have explored complementary mechanisms for evidence organization, contextual reasoning, and output verification. Data-centric generative systems can organize heterogeneous knowledge through semantic vector search, contextual querying, and agentic retrieval backends \cite{chen2026generative,liu2026diyhealth}.
Fine-grained evidence alignment and verification have also been studied through global-to-local semantic alignment \cite{zhu2024llms}, compact visual-evidence interfaces for intermediate reasoning \cite{gao2026visualthink}, and claim-level counter-evidence verification for hallucination detection and correction \cite{zhou2026hallucination}.
These studies collectively illustrate the broader value of explicitly organizing and validating evidence during model reasoning and generation.

\subsection{Patent Drafting}

While prior work has studied long-form generation in general domains, patent drafting introduces additional challenges due to strict structural, legal, and evidence-grounding requirements.
We next review patent drafting, covering both benchmark development and generation paradigms.

\textbf{Patent-related Benchmarks.} 
With the advancements of LLMs, the effort of testing their abilities in patent-related tasks has emerged.
Recent progress in LLMs has spurred benchmarking of patent-centric capabilities~\cite{jiang2025natural}.
For example, IPEval~\cite{ipeval} tests LLMs for patent law QA.
PatentEval~\cite{zuo2024patenteval} focuses on the tasks of claims-to-abstract and next-claim generations.
HUPD-DCG~\cite{hupd-dcg} measures patent claim generation performance from patent descriptions.
Complementary to task-specific benchmarks, Pap2Pat~\cite{pap2pat} extends evaluation to long-text patent drafting, reflecting growing interest in document-level generation. 
These benchmarks illustrate the evolution of patent-related evaluation from localized capabilities toward holistic long-form technical writing.

\vspace{-1mm}
\textbf{Patent-related Generation.}
Existing patent generation methods can be broadly grouped into three paradigms.
First is {\em patent-to-patent}, where a section of the patent content is generated from those of specific parts in the same patent.
For instance, PGT~\cite{christofidellis2022pgt} is pre-trained to support multiple patent-related tasks, e.g., title-to-abstract, abstract-to-claim.
PatentLMM~\cite{patentlmm} presents a large multimodal model to write descriptions of patent figures.
\cite{ren2025large} proposes a knowledge-tuning approach for initial concept discovery and content generation.
Second is {\em draft-to-patent}, where a draft of a patent is expanded and aligned with patent format and wording. 
PatentGPT~\cite{bai2024patentgpt} pre-trains a series of base LLMs and benchmarks them with patent generation from technical disclosure documents draft.
AutoPatent~\cite{wang2024autopatent} generates patents from draft information, where the draft is constructed through multiple rounds of iterative question–answering with an LLM under the guidance of a reference patent document.
Third is {\em paper-to-patent}, where an end-to-end process is studied from the original research paper to the patent draft.
For example, Pap2Pat~\cite{pap2pat} leverages ground-truth patent-derived outlines to guide the full-text patent migration.
However, prior methods either operate under partial-context settings or depend on externally provided (often manually crafted) drafting priors, leaving realistic long-form patent drafting with global logical organization less well addressed.





\section{Details on Semantic Chunking}
\label{sec:appendix-semantic-chunking}

To convert the raw content of the input document into knowledge units suitable for retrieval-augmented generation, we first perform semantic structure reconstruction based on page-level metadata extracted by PyMuPDF~\footnote{\url{https://pymupdf.readthedocs.io/}}, including textual blocks, font styles, coordinates, line spans, and paragraph layouts.
Using these layout cues, the system identifies the hierarchical boundaries of sections and paragraphs, thereby reconstructing the document's logical structure. 

Building upon this structure, we conduct semantic chunking to produce contextually coherent text segments that serve as the foundational elements of the vectorized document knowledge base. Unlike fixed-length segmentation methods~\cite{huang2023advancing,pap2pat}, our approach dynamically determines chunk boundaries according to the semantic coherence and contextual dependency between adjacent paragraphs.
Formally, given a document represented as an ordered sequence of paragraphs $\tilde{X} = \{\tilde{x}_1, \tilde{x}_2, \dots, \tilde{x}_m\}$, we compute the pairwise semantic similarity $s_{i,i+1}=\text{sim}(\mathcal{E}_p(\tilde{x}_i),\mathcal{E}_p(\tilde{x}_{i+1}))$, where $i=1,\cdots,m-1$, and $\mathcal{E}_p(\cdot)$ denotes the paragraph-level embedding function (e.g., Sentence-BERT~\cite{reimers2019sentence}).
The pairwise semantic similarity $\text{sim}(\cdot,\cdot)$ can be instantiated with different similarity metrics, while cosine similarity is adopted in this work due to its efficiency and empirical stability.
Paragraphs are iteratively merged when $s_{i,i+1} \geqslant \tau_s$ and the accumulated token length does not exceed the upper bound $t_{max}$.
This procedure yields a set of semantically coherent chunks $X = \{x_1, x_2, \dots, x_L\}$, where $L \leq m$ represents the total number of chunks for the input document and $x_i$ represents the $i$-th chunk aligned with the document’s logical hierarchy.
The resulting chunks constitute the minimal retrieval units for RAG, which are embedded into a high-dimensional semantic space and stored in the vector database as a vectorized document knowledge base, supporting structured querying, traceable grounding, and multi-turn semantic reasoning.

\section{Workflow of Logic Tree Construction}
\label{sec:appendix-workflow}

In this section, we provide the detailed algorithmic specification of the logic tree construction process used in \oursystem{}. As described in Section~\ref{sec:workflow}, logic tree construction constitutes the first stage of the overall patent generation pipeline, aiming to transform an unstructured research paper into a structured, evidence-grounded hierarchical representation. 
The procedure integrates semantic chunking, document-level vectorization, and evidence-guided recursive generation to progressively construct a logic tree that captures both the global structure and fine-grained technical details of the source document.
Algorithm~\ref{alg:logictree} presents the complete logic tree construction procedure.

\begin{algorithm}[t]
\caption{Logic Tree Construction}
\label{alg:logictree} 
\begin{algorithmic}[1]
\renewcommand{\COMMENT}[1]{\footnotesize {\color{gray}\texttt{#1}}}
\renewcommand{\algorithmicrequire}{\textbf{Input:}}  
\renewcommand{\algorithmicensure}{\textbf{Output:}}  
\REQUIRE Pre-publication research paper, the backbone model $\mathcal{M}_{LLM}$,
instructions $\sigma_I, \sigma_E, \sigma_R$,
token length bound $t_{max}$,
semantic chunking threshold $\tau_s$, discriminative relevance threshold $\tau_f$, content length budget $B$.
\ENSURE Logic tree $\mathcal{T}$.

\COMMENT{\scalebox{0.9}{$\setminus$*Semantic Chunking\&logic tree initialization*$\setminus$}}
\STATE Obtain an ordered sequence of paragraphs $X = \{x_1, x_2, \dots, x_m\}$ via layout-aware document analysis of the input paper
\FOR{$i=1$ {\bfseries to} $m-1$}
\STATE Compute the pairwise semantic similarity $s_{i,i+1}$ of $x_i$ and $x_{i+1}$ in $X$ using Eq.~\ref{eq:pairwise semantic similarity}
\IF{$s_{i,i+1} \geq \tau_s$ \textbf{and} $|x_i|+|x_{i+1}| \leq t_{max}$}
\STATE Merge $x_i$ and $x_{i+1}$ into a single segment
\ELSE
\STATE Retain $x_i$ and $x_{i+1}$ as separate segments
\ENDIF
\ENDFOR
\STATE Construct the final chunk set $\tilde{X} = \{\tilde{x}_1,\dots,\tilde{x}_L\},\, L\leq m$
\STATE Build vectorized document knowledge base $\mathcal{V}_{kb}=\{\mathcal{E}_p(\tilde{x}_i)\}_{i=1}^L$

\STATE Generate a draft technical description $\mathcal{D}_d=\mathcal{M}_{LLM}(\sigma_I,\tilde{X})$
\STATE Create root node $r$ of $\mathcal{T}$ with $c(r)=\mathcal{D}_d$ and $\mathcal{E}(r) = \tilde{X}$
\STATE Initialize the child nodes of $r$ by prompting $\mathcal{M}_{\text{LLM}}$ to decompose the key technical components of $\mathcal{D}_d$

\COMMENT{\scalebox{0.9}{$\setminus$*Evidence-Guided Recursive Generation with Node-Aware Semantic Search*$\setminus$}}
\STATE Initialize FIFO queue $\mathcal{Q}$ with all leaf nodes of $\mathcal{T}$
\REPEAT
\STATE Dequeue front node $v_t$ from $\mathcal{Q}$ with parent node $v_f$
\STATE Obtain candidate context spans $\tilde{\mathcal{D}}_{ctx}$ using Eq.~\ref{eq:candidate_context}
\STATE Calculate discriminative relevance
score $S_{dr}(d)$ for each $d\in \tilde{\mathcal{D}}_{ctx}$ using Eq.~\ref{eq:filtering_score}
\STATE Obtain context evidence $\mathcal{D}_{ctx}=\{d \,|\, S_{dr}(d)\geq \tau_f\}$

\IF{$\mathcal{D}_{ctx}$ is covered by $v\in \mathcal{T}\setminus V^{child}_f$}
\STATE Skip node $v_t$
\ELSE
\STATE Expand node $v_t$ using $\mathcal{M}_{LLM}$ with prompt $\sigma_E$ via Eq.~\ref{eq:expansion}
\STATE Update $\mathcal{T}$ and $\mathcal{Q}$
\IF{Under-specified term is detected from $c(v_t)$ using $\mathcal{M}_{LLM}$ with prompt $\sigma_{R1}\in \sigma_R$}
\STATE \scalebox{0.96}{Refine $c(v_t)$ using $\mathcal{M}_{LLM}$ with prompt in $\sigma_R$ via Eq.~\ref{eq:refinement}}
\ENDIF
\ENDIF
\UNTIL{$\mathcal{Q}$ is empty \textbf{or} $|\mathcal{T}|\geq B$}


\STATE {\bfseries Return} Logic tree $\mathcal{T}$
\end{algorithmic}
\end{algorithm}


\newpage
\section{Patent Generation Prompt}
\label{sec:appendix-prompt}

LogicTree-RAG employs a structured set of prompt instructions to control different stages of logic tree construction and recursive patent generation. Specifically, we design three categories of prompts: an initialization instruction $\sigma_I$, an expansion instruction $\sigma_E$, and a refinement instruction $\sigma_R$.
These prompts act as a lightweight control interface for the backbone LLM, enabling evidence-grounded, controllable long-form patent drafting.

The initialization instruction $\sigma_I$ is used to generate a draft technical description $\mathcal{D}_d$, which serves as the root node of the logic tree.
Given a set of semantically coherent chunks extracted exclusively from the introduction of the source paper, $\sigma_I$ prompts the model to synthesize a high-level technical description that captures the problem setting, core solution, and key system or method components, as detailed below.

\begin{lstlisting}[
style=promptstyle,
caption={Refinement instruction $\sigma_{R1}$ for under-specified term detection.},
captionpos=b,
label={lst:draft_prompt}]
@@@### ROLE@@@
You are an expert patent drafter and technical editor.

@@@### TASK DESCRIPTION@@@
Your task is to write a draft technical description of the invention, based ONLY on the provided document chunks of a research paper.

@@@### INPUT@@@
Input: a set of semantically coherent chunks derived exclusively from the paper introduction.

Document chunks:
"""
{CHUNKS}
"""

@@@### REQUIREMENTS@@@
Write a draft technical description that satisfies ALL requirements:
1. Capture the invention at a system and method level: clearly state (i) the technical problem, (ii) the core technical solution, and (iii) the key technical components. 
2. Make the description expansion-ready: include explicit technical handles such as module names/roles, input-output signals, key variables/parameters, and the end-to-end workflow, so that each part can be expanded later into detailed patent sections. 
3. Do not cite the paper, authors, or section numbers. Do not mention "chunks" or "retrieval".
4. Use ONLY information supported by the chunks.
5. No numbering, no bullet points, no headings.
6. Avoid vague hedges such as "may", "could", or "for example". Prefer definitive, technical phrasing.

Output ONLY the draft technical description text, with no extra commentary.
\end{lstlisting}


The expansion instruction $\sigma_E$ governs logic tree growth. Conditioned on a node contribution and its associated context evidence, $\sigma_E$ prompts the model to elaborate the contribution into finer-grained algorithmic details.
The expansion instruction $\sigma_E$ is specified as follows.

\begin{lstlisting}[
style=promptstyle,
caption={Expansion instruction $\sigma_E$ for \oursystem{} in long-form patent drafting.},
captionpos=b,
label={lst:expand_prompt}]
@@@### ROLE@@@
You are an expert in polishing patent applications, with professional knowledge of patent drafting and formatting standards.

@@@### TASK DESCRIPTION@@@
Your task is: given a specific claim contribution, extract the concrete algorithmic steps from the provided context and use them to expand the contribution, further specifying its algorithmic details.

@@@### REQUIREMENTS@@@
1. The output should consist of several paragraphs without numbering. The first paragraph must restate the given contribution point "{text}", and the subsequent paragraphs should describe the detailed algorithmic steps. Separate each paragraph with a blank line.  
2. The expansion must remain strictly focused on the content and concept described by the given contribution point, without introducing broader or unrelated concepts.  
3. The total output length should be longer than the input.  
4. When necessary, include key formulas and definitions. Do not insert line breaks inside formulas.  
5. Maintain precision and conciseness. Avoid terms such as "for example" or "may".
6. Each paragraph must be sufficiently detailed. If a step involves multiple sub-steps, describe them within a single paragraph using multiple sentences. The granularity of the steps should be consistent, with each step being necessary and indispensable.  

The original contribution point is: {TEXT}\n\n
The provided context is: {CONTEXT}\n\n
\end{lstlisting}

The refinement instruction set $\sigma_R=\{\sigma_{R1},\sigma_{R2},\sigma_{R3}\}$ addresses under-specified content during recursive generation.
Specifically, $\sigma_{R1}$ detects under-specified terms that lack sufficient technical detail, $\sigma_{R2}$ refines such terms using retrieved supporting evidence when available, and $\sigma_{R3}$ completes refinement using the model’s internal knowledge when no relevant evidence exists.
The refinement instructions $\sigma_R$ are instantiated as follows.

\begin{lstlisting}[
style=promptstyle,
caption={Refinement instruction $\sigma_{R1}$ for under-specified term detection.},
captionpos=b,
label={lst:refinement_prompt1}]
@@@### ROLE@@@
You are an expert patent drafter and technical editor.

@@@### TASK DESCRIPTION@@@
Your task is to identify under-specified terms in a patent specification node and propose how they should be refined.

@@@### DEFINITION@@@
Definition (under-specified term):
A term/phrase is under-specified if, for a technical disclosure, it lacks at least one of the following: a clear definition, a concrete algorithmic procedure, required inputs/outputs, parameters/hyperparameters, constraints/assumptions, or implementation-level details.

@@@### INPUT@@@
Input node content:
"""
{NODE_CONTENT}
"""

@@@### REQUIREMENTS@@@
1. Each "term" must be an exact substring of the input. 
2. Do not rewrite the node content. Do not add any extra keys. 
3. Prefer terms that are essential for technical completeness (e.g., key components, variables, algorithms, objectives, constraints).
4. Output ONLY the terms, one per line, with no numbering, no bullets, no extra text.
\end{lstlisting}

\begin{lstlisting}[
style=promptstyle,
caption={Refinement instruction $\sigma_{R2}$ for evidence-grounded targeted refinement.},
captionpos=b,
label={lst:refinement_prompt2}]
@@@### ROLE@@@
You are an expert patent drafter and technical editor. You will refine a specific under-specified term in a patent-specification node using the provided evidence.

@@@### TASK DESCRIPTION@@@
Your task is to produce an updated version of the node content that makes the target term "{TERM}" fully specified in a patent-appropriate and technically reproducible way.

@@@### INPUT@@@
Target term to refine:
"{TERM}"

Original node content:
"""
{NODE_CONTENT}
"""

Retrieved evidence (context spans):
"""
{EVIDENCE_SPANS}
"""

@@@### REQUIREMENTS@@@
1. All newly added technical details MUST be supported by the evidence. 
2. Preserve the original wording and structure as much as possible. Only add or revise the minimal necessary text to specify "{TERM}".
3. When applicable, specify inputs/outputs, key variables/parameters, computation steps, constraints/assumptions, and any formulas explicitly stated or implied by the evidence.
4. Use formal, precise language. Avoid hedges such as "may", "could", "for example". Avoid conversational tone.

Refined node content:
\end{lstlisting}

\begin{lstlisting}[
style=promptstyle,
caption={Refinement instruction $\sigma_{R3}$ for knowledge completion},
captionpos=b,
label={lst:refinement_prompt3}]
@@@### ROLE@@@
You are an expert patent drafter and technical editor. You will refine an under-specified term in a patent-specification node.

@@@### TASK DESCRIPTION@@@
Your task is to update the node content to make "{TERM}" sufficiently specified for a patent specification, using general and widely applicable technical descriptions.

@@@### INPUT@@@
Target term to refine:
"{TERM}"

Original node content:
"""
{NODE_CONTENT}
"""

@@@### REQUIREMENTS@@@
1. Preserve the original wording and structure as much as possible. Only add or revise the minimal necessary text to specify "{TERM}".
2. Provide a standard, implementation-oriented specification that would be reasonable across typical systems (e.g., define inputs/outputs, typical parameterization, and a generic procedure).
3. Use formal, precise language. Avoid hedges such as "may", "could", "for example". Avoid conversational tone.

Refined node content:
\end{lstlisting}


\section{Implementation Details}
\label{sec:appendix-implementation}

\subsection{Evaluation Metrics}
\label{sec:appendix-evaluation-metrics}
We evaluate patent generation performance from both content-level and language-level perspectives. Given the length and specialized nature of patent documents, automatic evaluation requires disentangling semantic content quality from linguistic realization. We therefore adopt established metrics that are well-suited for long-form technical text generation.

\noindent \textbf{Content-level Metrics.} Content-level metrics evaluate whether the generated patent conveys accurate and sufficient technical information grounded in the reference materials. All content-level metrics are computed at the document level following the evaluation protocol~\cite{pap2pat}.

\begin{itemize}[leftmargin=*]
\item \textbf{Coverage.}
Coverage measures the extent to which the generated patent covers the semantic content of the reference patent. We adopt the NLI-based SCALE metric~\cite{lattimer2023fast}, which estimates entailment between two long documents by aggregating sentence-level entailment scores.

Given a generated patent $X$ and a reference patent $R$, we treat $X$ as the \emph{premise document} and $R$ as the \emph{hypothesis document}. Due to the quadratic cost of exhaustive sentence-level comparisons, SCALE samples a set of hypothesis sentences $\{h_j\}_{j=1}^{N}$ from $R$. For each hypothesis sentence $h_j$, the top-$K$ most relevant premise chunks $\{p_{j,k}\}_{k=1}^{K}$ are retrieved from $X$ using BM25. An NLI model then computes entailment probabilities $P_{\text{entail}}(p_{j,k}, h_j)$.
The sentence-level entailment score is defined as:
\begin{equation}
s(h_j) = \max_{k} P_{\text{entail}}(p_{j,k}, h_j).
\end{equation}

The coverage score is computed by averaging over all sampled hypothesis sentences:
\begin{equation}
\text{Coverage}(X, R) = \frac{1}{N} \sum_{j=1}^{N} s(h_j).
\end{equation}

\item \textbf{Patent-grounded Factuality.}
Patent-grounded factuality ($\mathcal{F}_{Pat}$) measures the extent to which the generated patent is supported by the reference patent alone. In this setting, the reference patent $R$ serves as the premise document and the generated patent $X$ serves as the hypothesis document:
\begin{equation}
\mathcal{F}_{Pat}(X, R) = \text{SCALE}(R \Rightarrow G).
\end{equation}

\item \textbf{Source-grounded Factuality.}
Source-grounded factuality ($\mathcal{F}_{Src}$) measures semantic consistency when both the reference patent and the source paper are used as grounding evidence. Specifically, the premise document is formed by concatenating the reference patent $R$ and the source paper $P$, while the generated patent $X$ remains the hypothesis document:
\begin{equation}
\mathcal{F}_{Src}(X, R, P) = \text{SCALE}(R \cup P \Rightarrow G).
\end{equation}

This variant accounts for technically correct information present in the source paper but absent from the granted patent.

\item \textbf{ROUGE-L.}
ROUGE-L evaluates the longest common subsequence (LCS) between predicted and ground truth answers, reflecting recall-oriented similarity. It captures global structural overlap and complements BLEU's precision-based evaluation.
\begin{align}
\text{ROUGE-L} = 
\frac{(1 + \beta^2) \cdot R_{lcs} \cdot P_{lcs}}
{R_{lcs} + \beta^2 \cdot P_{lcs}},
\end{align}
where $R_{lcs} = \frac{\text{LCS}(X,Y)}{|Y|}, P_{lcs} = \frac{\text{LCS}(X,Y)}{|X|}.$ $Y$ is the ground truth patent, i.e., the reference patent, $X$ is the predicted text, i.e., the generated patent. $\text{LCS}(X,Y)$ is the length of the longest common subsequence. $\beta$ controls the weighting of recall versus precision.

\ignore{
\item \textbf{BERTScore.}
BERTScore measures semantic similarity between a generated text and a reference text using contextualized token embeddings. The final score is computed as the F1 measure:
\begin{equation}
\text{BERTScore} = \frac{2 \cdot \text{Precision} \cdot \text{Recall}}{\text{Precision} + \text{Recall}},
\end{equation}
where $\text{Precision} = \frac{TP}{TP + FP}, \text{Recall} = \frac{TP}{TP + FN}$. Here, $TP$, $FP$, and $FN$ are defined as soft quantities induced by token-level maximum cosine similarities between contextualized embeddings from BERT~\cite{devlin2019bert}, rather than discrete token matches.
}

\end{itemize}

\noindent \textbf{Language-level Metrics.}
Language-level metrics evaluate whether the generated patent adheres to stylistic conventions of patent writing, maintains discourse coherence, and avoids degenerate generation behaviors.

\begin{itemize}[leftmargin=*]
\item \textbf{Style.}
We measure stylistic similarity using corpus-wide n-gram profiles and StyloMetrix~\cite{okulska2023stylometrix}, as in~\cite{pap2pat}. For n-gram profiles, we construct frequency distributions over the 1{,}000 most frequent n-grams for $n \in \{1,2,3,4\}$ from the reference patents and the generated patents. Let $P^n_{\text{ref}}(g)$ and $P^n_{\text{gen}}(g)$ denote the normalized frequencies of n-gram $g$. The similarity score for each $n$ is computed as:
\begin{equation}
\text{sim}_n = \sum_{g \in P^n_{\text{ref}} \cup P^n_{\text{gen}}}
\left(
1 -
\left(
\frac{P^n_{\text{ref}}(g) - P^n_{\text{gen}}(g)}
{P^n_{\text{ref}}(g) + P^n_{\text{gen}}(g)}
\right)^2
\right).
\end{equation}

We additionally compute a StyloMetrix similarity score based on 196 rule-based linguistic features, including verb tense, modality, POS distributions, lexical patterns, and syntactic constructions. The final style score is computed as:
\begin{equation}
\text{Style} = \frac{1}{5}
\left(
\sum_{n=1}^{4} \text{sim}_n + \text{sim}_{\text{stylo}}
\right).
\end{equation}

\item \textbf{Repetition Rate.}
Repetition Rate ($\mathcal{R}_r$) is defined as the fraction of n-grams that appear more than once within a text~\cite{cettolo2014repetition}, given by:
\begin{equation}
\mathcal{R}_r =
\frac{\#\{\text{n-grams occurring more than once}\}}
{\#\{\text{total n-grams}\}}.
\end{equation}

To capture long-range repetition and degeneration effects, $\mathcal{R}_r$ is computed over sliding windows of 256 tokens and averaged across windows.

\item \textbf{Coherence.}
Discourse-level coherence is evaluated using DiscoScore~\cite{zhao2023discoscore}, a coherence metric based on Centering Theory~\cite{grosz1995centering} and contextualized representations from BERT~\cite{devlin2019bert}. DiscoScore assesses the consistency of entity transitions and discourse structure across adjacent sentences, producing a document-level coherence score.
Given a hypothesis document $X$, let $\mathbf{S}_{\text{gen}} \in \mathbb{R}^{n \times d}$ denote the matrix of sentence embeddings, where $n$ is the number of sentences and $d$ is the embedding dimension. A sentence graph $\mathcal{G}_{\text{gen}} = (V, A_{\text{gen}})$ is constructed, where each node corresponds to a sentence and the adjacency matrix $A_{\text{gen}}$ encodes discourse connectivity between sentence pairs. Sentence embeddings are updated via graph propagation $\hat{\mathbf{S}}_{\text{gen}} = (A_{\text{gen}} + I)\mathbf{S}_{\text{gen}}$, where $I$ is the identity matrix.

The same procedure is applied to the reference document $R$ to obtain $\hat{\mathbf{S}}_{\text{ref}}$. A document-level representation is then computed by aggregating sentence embeddings, yielding $\mathbf{G}_{\text{gen}}$ and $\mathbf{G}_{\text{ref}}$. The final DiscoScore is defined as the cosine similarity between the two graph-level representations:
\begin{equation}
\text{DiscoScore}(X,R) = \cos\!\left(\mathbf{G}_{\text{gen}}, \mathbf{G}_{\text{ref}}\right).
\end{equation}
\end{itemize}

Together, these metrics offer a comprehensive evaluation of patent generation quality that captures content accuracy, linguistic conventions, discourse coherence, and long-form generation behavior.

\subsection{Baseline Methods}
\label{sec:appendix-baselines}

For heuristic baselines, including the ground-truth patent and source paper,
no additional model inference is performed, and the documents are directly treated as generated outputs for metric computation.

Single LLM-call baselines generate the entire patent document in a single LLM invocation by conditioning on the full input paper using long-context prompting.
All baselines support 128k-token contexts, enabling full-paper ingestion and long-text drafting.
No intermediate structure or iterative refinement is employed.
Specifically, Kimi K2~\cite{team2025kimi} adopts a Mixture-of-Experts (MoE) architecture with 1T total parameters and 32B activated per token, and is pre-trained on 15.5T tokens with carefully engineered optimization strategies to ensure stable training at scale.
Qwen3-80B~\cite{abs-2505-09388} follows the Qwen3-Next design and integrates hybrid attention mechanisms and high-sparsity MoE layers, enabling efficient context modeling under ultra-long context lengths while maintaining computational efficiency.
Qwen3-Max~\cite{abs-2505-09388} is a proprietary flagship model exceeding one trillion parameters, pre-trained on 36T tokens, and inherits the MoE-based design of the Qwen3 series with enhanced load balancing and stability-oriented training strategies.
DeepseekV3~\cite{abs-2412-19437} is an MoE model with 671B total parameters and 37B activated per token, combining Multi-head Latent Attention and DeepSeekMoE architectures, and is pre-trained on 14.8T tokens with multi-token prediction objectives to support efficient and stable large-scale training.
GPT-5~\cite{achiam2023gpt} and Gemini-2.5 Pro~\cite{comanici2025gemini} are proprietary frontier LLMs evaluated under deep thinking mode, which enables stronger deliberative reasoning and more careful long-context synthesis. 
They serve as strong closed-source baselines for assessing whether advanced reasoning-oriented prompting alone can produce coherent and complete long-form patent drafts without explicit structural orchestration.
Collectively, these models provide strong capacity for long-context representation and serve as competitive backbone LLMs for single LLM-call patent generation.
Unless otherwise specified, these baselines use the same decoding configurations as LogicTree-RAG.

Furthermore, we include Chunk-based Outline-guided Patent Generation (COPGEN)~\cite{pap2pat} and its supervised fine-tuned (SFT) variant as representative multi-stage LLM-based patent generation methods.
Specifically, COPGEN chunks the outline, selects the most
relevant parts of the source paper depending on the chunks' outline bullet points, and prompts an LLM to generate the patent text for that chunk.
To ensure a fair comparison under our problem setting, where no ground-truth patent outline is available at inference time, we report two variants of COPGEN.
The first follows the outline-empty configuration, where only section titles are provided.
For completeness, we also report results obtained using the long-version outlines provided by the Pap2Pat benchmark.
For the SFT variant, only the backbone LLM is fine-tuned on the Pap2Pat training set (1,000 paper-patent pairs), and no test data is used during training.

In addition, we consider recent long-form generation frameworks LongWriter~\cite{bai2025longwriter} and LongWriter-Zero~\cite{wu2026longwriter}, which decompose long-form generation into multi-stage processes to improve global coherence and content planning. 
We also include general-purpose agentic paradigms, including ReAct~\cite{yao2023react} and Plan-then-Execute~\cite{he2025plan}, which enable LLMs to perform step-wise planning and tool use without relying on hard-coded structural templates.
To adapt them to our setting, we expose our core components, i.e., semantic chunking, node-aware retrieval, and evidence-guided generation, as callable tools, and replace the logic-tree-based orchestration with generic agent planning strategies.

\subsection{Hyperparameters and Decoding Configurations}
\label{sec:appendix-Hyperparameters}

In this subsection, we summarize the hyperparameters and decoding configurations adopted for \oursystem{} and the baseline methods.
\oursystem{} adopts Qwen3-80B~\cite{abs-2505-09388} as the backbone LLM throughout all experiments, unless stated otherwise.
We set the semantic chunking threshold $\tau_s$ to 0.8, and the accumulated token length $t_{max}$ to 512.
For node-aware semantic search, we first retrieve up to 100 candidate evidence spans $\tilde{\mathcal{D}}_{ctx}$ from the document
knowledge base $\mathcal{V}_{kb}$.
These candidates are then re-ranked using the discriminative relevance score.
Rather than using a fixed score threshold, we retain the top-5 candidates and set the discriminative relevance threshold $\tau_f$ adaptively as the score of the 5-th ranked candidate.
To ensure controllable expansion during evidence-guided recursive generation, we impose a content length budget $B$ that limits the maximum size of the logic tree.
Specifically, the recursive process terminates when the number of nodes in the tree reaches 64, preventing unbounded expansion while preserving sufficient structural coverage.
In addition, we constrain the maximum depth of the logic tree to 3 (with the root node at depth 0), which avoids overly deep recursive expansion and encourages a balanced hierarchical structure. 
Unless otherwise specified, these values are fixed across all experiments.

Reported results in Section~\ref{sec:exps-main} are evaluated on the Pap2Pat test set, which consists of 500 paper-patent pairs.
For the SFT variant of COPGEN, fine-tuning is conducted exclusively on the Pap2Pat training set containing 1,000 pairs.
All LLM-based methods use identical decoding configurations, with a temperature of 0.3, top-$p$ of 1.0, and a maximum generation length of 8,192 tokens.
The SFT variant of COPGEN fine-tunes the backbone LLM Llama-3 8B~\cite{dubey2024llama} using LoRA~\cite{hu2022lora}.
Following~\cite{pap2pat}, we adopt a maximum sequence length of 8,192 tokens and train the model for 3 epochs with a batch size of 32.
The learning rate is set to $3.16\times10^{-4}$ with a cosine scheduler and a warmup ratio of 0.1.
For LoRA-specific configurations, we use a rank of 128, a scaling factor ($\alpha$) of 60, and a dropout rate of 0.05.

\subsection{Experimental Environment}
\label{sec:appendix-Environment}

All experiments are conducted on a server equipped with an AMD EPYC 9334 CPU @ 2.7 GHz (32 cores), 128 GB of memory, and an NVIDIA H100 NVL with CUDA 12.9.
The operating system is Ubuntu 24.04 running Linux kernel 6.8.0-79-generic.
LLMs used in our experiments are accessed via standardized cloud-based inference APIs provided by Alibaba Cloud Bailian.

\subsection{Detailed Prompt Settings}
\label{sec:appendix-prompt-settings}

This subsection details the prompt configurations used in our experiments. For COPGEN and its variants, we adopt the prompt settings from the publicly released implementation\footnote{\url{https://github.com/boschresearch/Pap2Pat}}. For single LLM-call baselines, we employ a unified prompt template following COPGEN to ensure a fair and consistent comparison, as described below.


\begin{lstlisting}[
style=promptstyle,
caption={System prompt for single LLM-call baselines in long-form patent drafting.},
captionpos=b,
label={lst:baseline_prompt_sys}]
@@@### ROLE@@@
You are a highly skilled patent attorney with decades of experience in drafting high-quality patent applications.
You assist scientists in transforming their scientific discoveries into lucrative patents.

@@@### TASK DESCRIPTION@@@
Your task is to draft a patent application.

@@@### INPUTS@@@
As input, you will be provided a research paper and a patent outline, each serving a distinct purpose.

1. Research Paper:
The research paper describes a novel invention to be patented. Your task is to extract the invention from the paper and write a patent application for it.

2. Patent Outline:
The patent outline summarizes the desired discourse structure of the patent document. Use this outline as a rough guidance during drafting.

@@@### GUIDELINES@@@
* Copy the headings from the outline exactly. You must include only the headings provided in the outline.
* You must always write complete sentences and avoid keywords, bullet lists and enumerations.
* The patent must act as a standalone document, therefore do not refer to the research paper in the patent.

\end{lstlisting}

\newpage
\begin{lstlisting}[
style=promptstyle,
caption={User prompt for single LLM-call baselines in long-form patent drafting.},
captionpos=b,
label={lst:baseline_prompt_user}]
Here is the outline of the desired patent application.
Per bullet point, write roughly 4000 words.

@@@### DESCRIPTION OF THE INVENTION@@@
In the example above, each line beginning with '#' is a bullet point.
```md
# DESCRIPTION

## FIELD OF THE INVENTION

## BACKGROUND OF THE INVENTION

## SUMMARY OF THE INVENTION

## DETAILED DESCRIPTION
```
You need to draft a complete patent application that strictly follows the outline's section order and headings.
Do not skip any bullet points. Use formal patent language. Remember what you have already written and do not repeat yourself.

@@@### RESEARCH PAPER@@@
Here is the research paper that describes the invention:
```
# Introduction

Safe delivery of a highly conformal dose distribution to a well-defined target volume in radiotherapy is not an easy task. It has become increasingly challenging due to the advent of advanced treatment techniques such as intensity-modulated radiation therapy (IMRT) and volumetric-modulated arc therapy (VMAT). Though a downward trend in radiotherapy incident rates has been indicated by several reports (1, 2), severe incidents with detrimental effects, including death, have been reported recently and received public attention. (3)  Radiotherapy is a complicated, multistep, multiperson process, and errors can occur at any point. One of the prominent causes for radiotherapy incidents is geometric miss due to incorrect patient setup, which leads to the treatment of incorrect body parts with more than 1 cm spatial discrepancy (3, 4). Geometric miss can result in significant underdose to the target, which can cause tumor recurrence, and overdose to healthy tissue with severe normal tissue complications. In hypofractionated radiotherapy, such as stereotactic body radiotherapy (SBRT), it may result in even more severe morbidity than traditional radiotherapy.

Several reports have identified geometric miss caused by incorrect patient setup as the leading cause of radiotherapy errors. In a study of 100 radiotherapy incidents reported internationally, (3) the frequency of incidents due to incorrect patient setup was reported to be 21%. Excluding the 44 incidents involving brachytherapy, in which patient setup accuracy was less of an issue, this frequency would rise significantly to 37.5%. (3)  Clark et al. (2)  analyzed clinical incidents reported internally within a large academic center between 2007 and 2009, and reported that 14 out of 41 critical, major, or serious incidents were geographic misses. Among the 14 incidents, one was caused by wrong target identification in treatment planning, while the other 13 incidents were due to shifting errors at patient setup. In an online report analyzing event causes for 230 misadministrations reported between 2001 and 2009 in New York from the New York State Department of Health, incorrect body part treated was the leading type of errors at 46%, most often due to incorrect patient setup. (5)

The challenge in patient setup is to accurately localize the patient to the same treatment position as planned in each treatment session. Over the past decade, there has been a rapid expansion in technological tools to facilitate accurate patient localization. Examples include immobilization devices with couch indexing capability and image-guided radiotherapy (IGRT). Indexed immobilization devices not only ensure repeatable patient fixation but also reduce patient setup errors by providing initial approximate target localization. X-ray-based image-guided technologies, such as cone-beam CT (CBCT), enable visualization of internal anatomy with sufficient soft tissue contrast and allow corrections for misalignment or interfraction motion through registration with reference CT images (6). IGRT systems capable of continuous tracking, such as AlignRT (Vision RT Ltd., London, UK), C-Rad Sentinel (C-Rad AB, Uppsala, Sweden), SonArray (Varian Medical Systems, Palo Alto, CA), and ExacTrac Optical Tracking System (BrainLAB, Heimstetten, Germany), have also been developed for patient setup guidance. These systems have the potential to eliminate patient setup errors and significantly reduce setup uncertainty, as demonstrated by multiple studies. Bissonnette and Medlam (1) reported a 50 percent overall decrease in incidents caused by localization errors following the widespread introduction of IGRT on six of their 16 linear accelerators. Clark et al. (2) noted that none of the 13 geometric misses occurred on treatment units equipped with daily image guidance systems.

However, the use of advanced technological equipment does not guarantee that radiotherapy is immune to setup errors (7, 8). Patient setup is a process that cannot be fully automated and is therefore subject to human error. Several contributing factors have been identified. First, sufficient formal training is not always provided to personnel who operate the devices and interpret the results. Incorrect interpretation can lead to improper adjustment of treatment position and consequently to setup errors. For example, when treating the thoracic spine, there is a significant risk of targeting the wrong vertebral body (9). Due to similarities in bony anatomy in this region, incorrect alignment may occur as a result of misregistration using orthogonal imaging or CBCT. Second, as the complexity of treatment tools increases, the complexity of device control and the workload for therapists also increase, as reflected by the growing number of computer monitors in control rooms (8). In the absence of streamlined workflows and standardized controls, therapists may lose attention to the correctness of treatment delivery. Third, although most IGRT devices provide high geometric precision (10), they have inherent limitations that may compromise patient safety. For example, commonly used radiographic systems such as CBCT do not track patient position changes and therefore represent patient position only at the time of image acquisition. After CBCT imaging, the treatment couch may be moved for specific reasons, such as clearance checks, and not returned to the intended treatment position. Modern delivery systems include interlocks to detect such errors but also allow user overrides. If an interlock is inappropriately overridden, the patient may be treated at an incorrect site. In addition, CBCT and most radiographic-based IGRT systems cannot be used for noncoplanar setups, resulting in discrepancies between imaging and treatment positions when noncoplanar beams are applied (11). Continuous tracking systems such as ExacTrac Optical Tracking System, AlignRT, and C-Rad Sentinel are typically employed only for limited disease sites in current clinical practice. Therapists operating machines equipped with multiple IGRT systems often switch between devices depending on the treatment site, further increasing operational complexity. Fourth and most importantly, immediate and independent position verification is not always available to confirm that the patient has been positioned exactly as planned, particularly when the limitations of the employed IGRT systems are encountered (8).

In this paper, we present an efficient automatic patient safety system (PSS) that addresses factors contributing to gross setup errors and safeguards patient treatment. The system uses CCD cameras to track a single infrared reflective marker affixed to the patient's skin or immobilization device. It is a general-purpose system applicable to all disease sites and provides continuous and independent verification of patient setup accuracy. With a fully automated workflow, the PSS serves as an effective complement to existing IGRT systems to enhance patient safety in radiotherapy.

# Materials And Methods

## A. System Overview

The system consists of a pair of Polaris CCD cameras (Northern Digital Inc., Waterloo, Ontario) mounted on the treatment room ceiling (see Fig. 1(a)), a single IRRM affixed to the patient's skin or immobilization device, and a set of in-house developed software. The software communicates with the control box of the cameras through a standard serial cable and a proprietary data cable. An internal processing unit performs triangulation to determine the coordinates of IRRMs in its native coordinate system. The three-dimensional (3D) coordinates of up to 50 IRRMs can be transmitted to the computer simultaneously. System calibration, workflow, and daily quality assurance (QA) procedures are described in the following sections.

## B. System Calibration

A calibration procedure was developed to convert the camera's native coordinate system to the absolute room coordinate system. A calibration jig consisting of five commercial IRRMs (Fig. 1(b)) was used for this purpose. The coordinates of each IRRM relative to the center of the jig were determined via CT scan and collected in a 5x3 matrix A. Under the guidance of CBCT, the jig was placed on the treatment couch with its center aligned with the imaging isocenter. The congruence of the imaging isocenter and the machine isocenter was confirmed using a ball bearing phantom (6). The coordinates of the IRRMs were sampled by the cameras and collected into another matrix B. A rotation matrix R and a translation matrix S were estimated by minimizing the residual error between the transformed and reference coordinates.

This is known as the relative pose problem in computer vision and has been studied extensively. 12,  13,  14  In this work, we employed a general analytical solution via singular value decomposition (SVD), (14)  and the steps were summarized as follows: The centroids (Ac) and (Bc) were subtracted from A and B, respectively, to align the origins of the two coordinate systems. The SVD of BT A was computed as \(B^{T}A = \text{UWV}^{T}\) where U and V are unitary matrices, and W is a diagonal matrix whose diagonal entries are equal to the singular values of BT A.

3. The solutions of R and S were given as \(\begin{matrix} {R = U\begin{pmatrix} 1 & 0 & 0 \\ 0 & 1 & 0 \\ 0 & 0 & {\text{det}(UV^{T})} \end{pmatrix}V^{T}} \\ {\quad\quad\quad\quad\quad\quad\quad\quad\quad\quad\quad\quad\text{and},} \\ {S = A_{c} - B_{c}R} \end{matrix}\)

## System Workflow

Great emphasis was placed on developing a smooth clinical workflow for the system. The workload added to the therapists should be kept to a minimum, which was critical for the system to be adopted into clinic use. A minimal workload would also help minimize the possibilities of human errors. The workflow of the PSS is described in the following sections as the preparation stage and the treatment stage. Figure 2 shows a demonstrative workflow of the system when used on head and neck (H&N) patients treated with aquaplastic face masks.

...

# Conclusions

We have developed an efficient automatic general-purpose PSS to prevent gross setup errors in radiotherapy. The system provides real-time independent position verification for the treatment of all disease sites based on optical-tracking technology and is independent of treatment room, treatment machine, couch top design, immobilization device, or patient internal anatomy. The system has been well adopted for use in a busy clinical environment because of its seamless workflow and effectiveness in catching setup errors. Due to its advantages of continuous tracking capability, no radiation dose, and fully automated clinical workflow, it serves as an ideal complement to complex IGRT systems in ensuring patient safety in radiotherapy.
```
\end{lstlisting}


\section{Supplementary Experimental Results}
\label{sec:appendix-exps}






\subsection{Supplementary Ablation Study}
\label{sec:appendix-ablation}

In this subsection, we evaluate the robustness of \oursystem{} by instantiating the framework with different backbone LLMs, while keeping all other components and hyperparameters unchanged. As shown in Table~\ref{tab:exp-backbone}, \oursystem{} consistently produces long and well-structured patent drafts across all backbones, indicating that controlling the tree size enables stable and sufficient output length independent of the underlying model.
Content-level metrics such as coverage and factuality remain stable, while language-level metrics, including style and repetition rate, gradually improve with stronger backbones. Notably, when both \oursystem{} and COPGEN~\cite{pap2pat} are instantiated with the same Llama-3-8B backbone, \oursystem{} achieves substantially better performance and approaches the performance of the SFT-enhanced COPGEN variant, highlighting the effectiveness of the proposed logic-centric generation framework without relying on task-specific fine-tuning.
Overall, these results demonstrate that the gains of \oursystem{} stem primarily from its structured generation paradigm rather than dependence on a particular backbone model.

\begin{table}[ht]
   \small
    \centering
    \renewcommand{\arraystretch}{1.5}
    \caption{Performance of \oursystem{} with different backbone LLMs.
    }
    \label{tab:exp-backbone}
    \resizebox{1\columnwidth}{!}{
        \begin{tabular}{l ccccc ccc c}
\toprule
 \multirow{3}{*}{Method} & \multirow{3}{*}{\makecell[c]{Output\\Tokens}} & \multicolumn{4}{c}{Content-level} & \multicolumn{3}{c}{Language-level}  & \multirow{3}{*}{\makecell[c]{Time (s)\\ / Patent (Token)}} \\
 \cmidrule(lr){3-6} \cmidrule(lr){7-9}
& & \multirow{2}{*}{Coverage $\uparrow$} & \multicolumn{2}{c}{Factuality $\uparrow$} & \multirow{2}{*}{\makecell[c]{Semantic \\ $\,\,$ Similarity $\uparrow \,\,$}} & \multirow{2}{*}{Style $\uparrow$} & \multirow{2}{*}{\makecell[c]{Repetition Rate\\$\mathcal{R}_r$ $\downarrow$}} & \multirow{2}{*}{Coherence $\uparrow$}\\
\cmidrule(lr){4-5} 
& & & $\mathcal{F}_{Pat}$ & $\mathcal{F}_{Src}$ &  & &  & \\
  \toprule[0.8pt]
\multicolumn{10}{c}{Heuristic Skyline}\\
Reference Patent (GT)  & 18.16k & 89.05 & 88.74 & 88.80 & 100 & 100 & 13.91 & 100 & - \\
\hdashline
Source Paper & 8.03k & 44.77 & 46.06 & 88.68 & 39.37 & 39.26 & 8.39 & 98.60 & -\\
\hdashline
Outline (Long) & 1.43k & 39.11 & 61.79 & 61.93 & 10.16 & 23.22 & 20.14 & 85.69 & -\\
\midrule
\multicolumn{10}{c}{\oursystem{}} \\

Llama3-8B~\cite{dubey2024llama} & 16.25k & 40.25 & 56.70 & 61.95 & 35.74 & 55.89 & 7.16 & 96.98 & 209 (12.86) \\
\hdashline
Qwen2.5-72B~\cite{team2024qwen2} &17.82k & 42.08 & 57.42 & 66.09 & 34.89 & 62.28 & 5.87 & 97.56 & 260 (14.58) \\
\hdashline
DeepseekV3~\cite{abs-2412-19437} & 19.15k & 43.20 & 58.21 & 64.83 & 37.95 & 58.24 & 5.86 & 97.10 & 320 (16.69)\\
\hdashline
Qwen3-80B~\cite{abs-2505-09388} & 20.22k & 43.79 & 58.92  & 66.34 & 38.91 & 65.68 & 4.14 & 97.45 & 311 (15.38) \\

\bottomrule
        \end{tabular}
        }
\end{table}

\subsection{Supplementary Study on Agentic Planning}
\label{sec:appendix-agentic}

To directly evaluate the role of structured planning, we compare LogicTree-RAG with two representative agentic paradigms, ReAct~\cite{yao2023react} and Plan-then-Execute~\cite{he2025plan}.
To ensure a fair comparison, we expose our core modules (i.e., semantic chunking, node-aware retrieval, and evidence-guided generation) as callable tools, and replace the logic-tree-based orchestration with generic agent planning strategies. This design isolates the effect of explicit structural planning (logic tree) versus generic agentic decomposition, while keeping underlying capabilities comparable.

As shown in Table~\ref{tab:exp-agentic}, despite using the same underlying tools, the agentic variants exhibit lower coverage, weaker factual grounding, and higher repetition, indicating incomplete and less controlled generation. In contrast, LogicTree-RAG achieves more balanced and consistent outputs, with superior performance across both content-level and language-level metrics.
These results suggest that generic agentic planning alone is insufficient for long-form structured generation. Instead, explicit global structure modeling via the logic tree is critical for coordinating content organization, evidence grounding, and generation control.

\begin{table}[ht]
   \small
    \centering
    \renewcommand{\arraystretch}{1.2}
    \caption{Performance comparison of LogicTree-RAG against agentic baselines for patent drafting.
    }
    \label{tab:exp-agentic}

    \resizebox{1\columnwidth}{!}{
        \begin{tabular}{l ccccc ccc c}
\toprule
\tikzmark{tl} \multirow{3}{*}{Method} & \multirow{3}{*}{\makecell[c]{Output\\Tokens}} & \multicolumn{4}{c}{Content-level} & \multicolumn{3}{c}{Language-level}  & \multirow{3}{*}{\makecell[c]{Time (s)\\ / Patent (Token)}} \\
 \cmidrule(lr){3-6} \cmidrule(lr){7-9}
& & \multirow{2}{*}{Coverage $\uparrow$} & \multicolumn{2}{c}{Factuality $\uparrow$} & \multirow{2}{*}{\makecell[c]{Semantic \\ $\,\,$ Similarity $\uparrow \,\,$}} & \multirow{2}{*}{Style $\uparrow$} & \multirow{2}{*}{\makecell[c]{Repetition Rate\\$\mathcal{R}_r$ $\downarrow$}} & \multirow{2}{*}{Coherence $\uparrow$}\\
\cmidrule(lr){4-5} 
& & & $\mathcal{F}_{Pat}$ & $\mathcal{F}_{Src}$ & & &  & \tikzmark{br}\\
  \toprule[0.8pt]

LogicTree-RAG & \textbf{20.22k} & \textbf{43.79} & \textbf{58.92}  & \textbf{66.34} & \textbf{38.91} & \textbf{65.68} & \textbf{4.14} & \textbf{97.45} & 311 (15.38) \\
\newarrow w/ ReAct   & 16.86k & 40.28 & 55.98 & 64.85 & 36.86 & 65.52 & 10.24 & 96.18 & 336 (19.93)\\
\newarrow w/ Plan-Execute & 18.58k & 42.02 & 56.35 & 64.96 & 35.91 & 64.97 & 8.95 & 96.59 & 325 (17.49)\\
\bottomrule
        \end{tabular}
        }
\end{table}


\subsection{Evidence-support Analysis}
\label{sec:appendix-evidence}
To assess retrieval-grounded generation at the evidence level, we evaluate whether generated content is supported by retrieved evidence for each retrieval-triggered node, using an LLM-as-judge (GPT-5) and human audit. Each case is categorized as fully supported, partially supported, or unsupported. We further conduct human auditing on 20 randomly sampled test cases under the same criteria. Results are summarized in Table~\ref{tab:exp-evidence}.
LogicTree-RAG achieves a substantially higher proportion of fully supported content and fewer unsupported cases than COPGEN, indicating stronger attribution faithfulness and more effective use of retrieved evidence. Human auditing yields consistent trends, confirming that retrieved evidence is actively used to support generation rather than being passively retrieved.
These findings align with the ablation results in Table~\ref{tab:component-wise-ablation}, where removing evidence-grounded refinement $\sigma_{R2}$ consistently degrades performance, while removing internal refinement $\sigma_{R3}$ has a negligible impact. These results suggest that the gains of LogicTree-RAG primarily arise from improved evidence-grounded generation, rather than increased pipeline complexity or auxiliary mechanisms.

\begin{table}[ht]
\centering
\renewcommand{\arraystretch}{1.1}
\caption{Evidence-support analysis of retrieved evidence and generated content.
}
\label{tab:exp-evidence}
\resizebox{0.6\columnwidth}{!}{
\begin{tabular}{lccc}
\toprule
Method  &  Fully (\%) & Partial (\%) & Unsupported (\%) \\
\midrule
\multicolumn{4}{c}{LLM-as-judge}\\
\hdashline
COPGEN & 70.4 & 20.2 & 9.4 \\
LogicTree-RAG & 85.6 & 12.0 & 2.4 \\
\midrule
\multicolumn{4}{c}{Human Audit}\\
\hdashline
COPGEN & 60 & 30 & 10 \\
LogicTree-RAG & 75 & 25 & 0 \\
\bottomrule
\end{tabular}
}
\end{table}

\subsection{Computational Cost Analysis}
\label{sec:appendix-cost}

To improve transparency, we report a detailed cost breakdown of LogicTree-RAG along the workflow pipeline. As shown in Table~\ref{tab:appendix-cost}, the overall cost is well controlled, with a balanced distribution across stages. In particular, recursive logic tree construction introduces only moderate overhead, while the majority of cost arises from traversal and generation.
Importantly, the recursion depth is explicitly bounded by tree budget hyperparameters (see Appendix~\ref{sec:appendix-Hyperparameters}), preventing uncontrolled expansion and ensuring a favorable quality-efficiency trade-off rather than gains driven by excessive API usage.

\begin{table}[ht]
\centering
\renewcommand{\arraystretch}{1.2}
\caption{Supplementary computational cost analysis.}
\label{tab:appendix-cost}
\resizebox{0.8\columnwidth}{!}{
\begin{tabular}{ll}
\toprule
Computational Cost (/patent) & Cost Breakdown\\
\midrule
LLM Calls: 20 & chunking\&tree init 5\%, tree construction 45\%, traversal 50\% \\

Runtime: 311s & chunking\&tree init 16\%, tree construction 39\%, traversal 45\% \\

Total Input Tokens: 7.60k & chunking\&tree init 29\%, tree construction 37\%, traversal 34\% \\

\bottomrule
\end{tabular}
}
\end{table}

\subsection{Human Expert Study}
\label{sec:appendix-expert}

We conduct a human expert evaluation with three patent practitioners to assess the practical quality of generated patents. We randomly select 10 samples from the test set.
For each sample, practitioners are provided with the source paper, the generated patent, and the reference patent, along with the corresponding outlines (long version) to ensure fair comparison.
Experts perform pairwise comparisons (win/loss/tie) between \oursystem{} and each baseline. We adopt five evaluation criteria covering both legal correctness and technical quality, as defined in Listing~\ref{lst:GUIDELINES}. All evaluations are conducted in a blinded manner, where the identities of methods are anonymized.

The win-rate results are summarized in Table~\ref{tab:exp-human}. \oursystem{} consistently outperforms all baselines across evaluation criteria, achieving strong human preference in overall quality and technical correctness.
In particular, \oursystem{} shows substantial advantages in Claim Structure and Antecedent Basis, indicating improved structural coherence and consistency in long-form patent drafting. Gains in Fidelity further demonstrate that the generated patents remain closely grounded in the source content.

\begin{lstlisting}[
style=promptstyle,
caption={Definitions of evaluation criteria used in human expert assessment.},
captionpos=b,
label={lst:GUIDELINES}]
@@@### Expert Evaluation Criteria@@@
* Overall Quality: overall usefulness as a patent draft for practitioner revision
* Legality: conformity to basic patent drafting and legal-form requirements
* Claim Structure: clarity and correctness of claim organization and dependencies
* Antecedent Basis: consistency of terminology and antecedent basis
* Fidelity: faithfulness to the technical content of the source paper
\end{lstlisting}

\begin{table}[ht]
\centering
\renewcommand{\arraystretch}{1.1}
\caption{Win-rate results of LogicTree-RAG in human expert evaluation.
}
\label{tab:exp-human}
\begin{tabular}{lccccc}
\toprule
Method & Overall & Legality & Claim Structure & Antecedent Basis & Fidelity \\
\midrule
Ours vs Qwen3-80B & 75.7 & 69.8 & 84.6 & 84.8 & 88.5 \\
Ours vs DeepseekV3 & 84.4 & 65.1 & 88.5 & 90.1 & 90.2 \\
Ours vs LongWriter & 78.8 & 76.5 & 90.6 & 92.0 & 85.9 \\
Ours vs COPGEN & 70.5 & 65.7 & 85.2  & 82.1 & 78.5\\
\bottomrule
\end{tabular}
\end{table}

\subsection{Sensitivity to Tree Growth Budget}
\label{sec:appendix-treesize}
\begin{figure}[ht]
\begin{center}
\centerline{\includegraphics[width=0.6\linewidth]{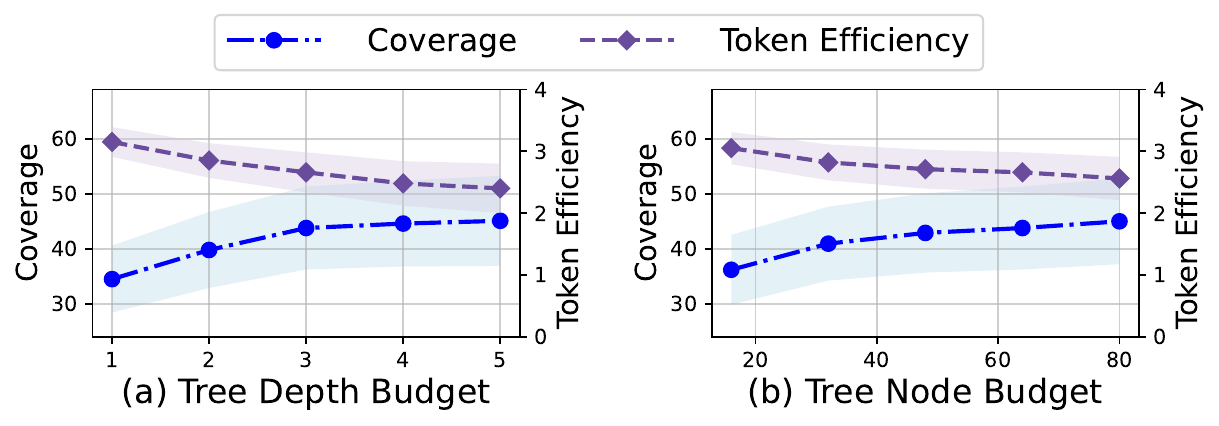}}
\caption{Sensitivity to tree growth budget in \oursystem{}.}
\label{fig:exp-sensitivity-tree}
\end{center}
\end{figure}

We further analyze the sensitivity of \oursystem{} to the tree growth budget by varying the maximum tree depth and the maximum number of tree nodes, while keeping all other components and hyperparameters fixed.
As shown in Figure~\ref{fig:exp-sensitivity-tree}, increasing the tree budget consistently improves coverage, indicating that deeper or broader trees enable more comprehensive exploration of the document structure. However, the gains diminish beyond moderate budgets, suggesting that additional expansions primarily capture long-tail details.
In particular, the default configuration, i.e., maximum depth = 3 and maximum number of nodes = 64, lies in this favorable regime, achieving high coverage with reasonable efficiency.
In contrast, token efficiency decreases as the tree budget grows, due to the increased retrieval and generation overhead incurred by additional nodes.
These results reveal a clear quality–cost trade-off and demonstrate that \oursystem{} allows controllable scaling, where moderate tree budgets achieve a favorable balance between coverage and efficiency.

\newpage
\subsection{Case Study}
\label{sec:appendix-case}
\begin{figure}[ht]
\begin{center}
\centerline{\includegraphics[width=\linewidth]{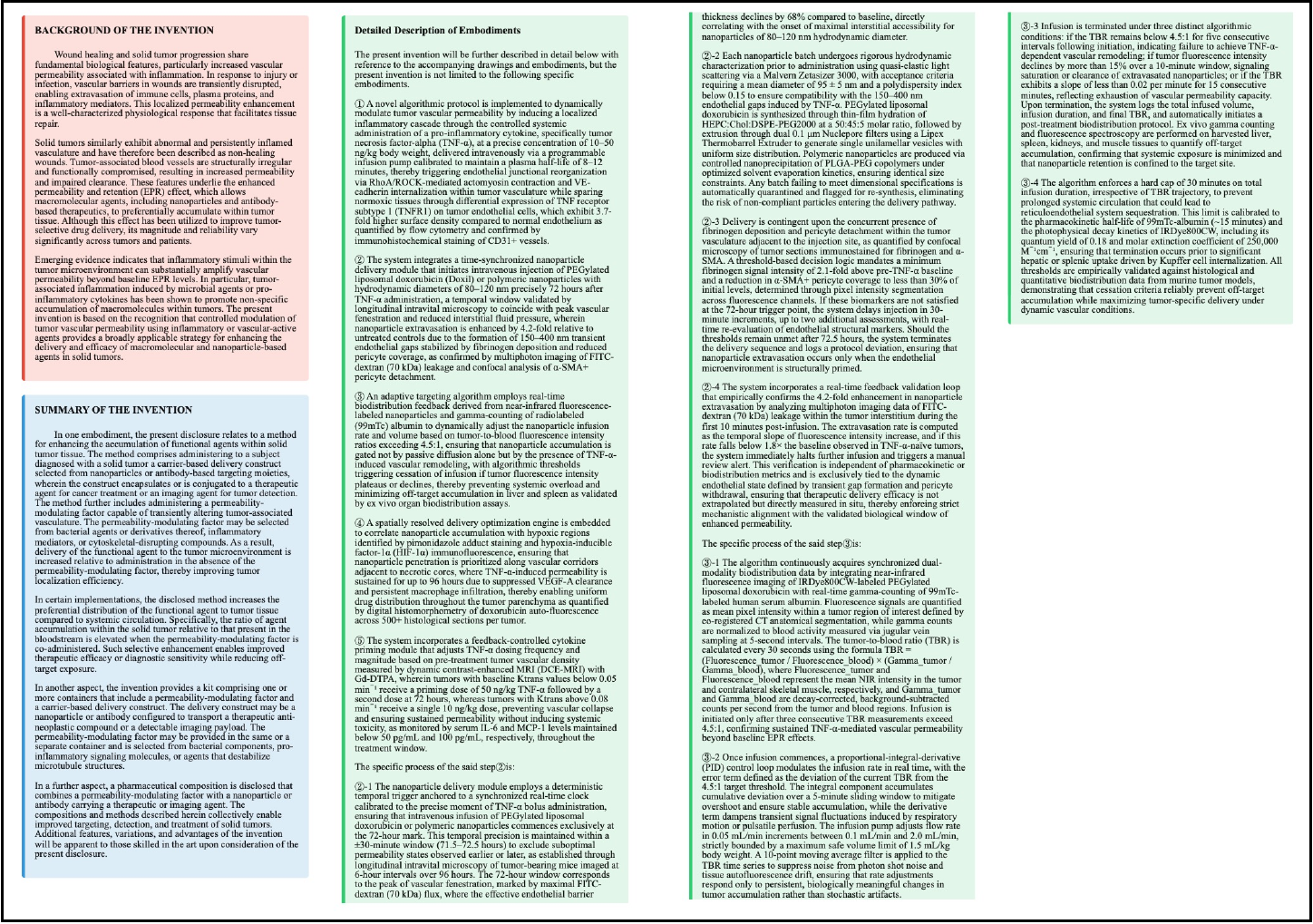}}
\caption{Qualitative case study illustrating a patent example generated by \oursystem{}.}
\label{fig:case_study}
\end{center}
\end{figure}

\begin{figure}[t]
\begin{center}
\centerline{\includegraphics[width=\linewidth]{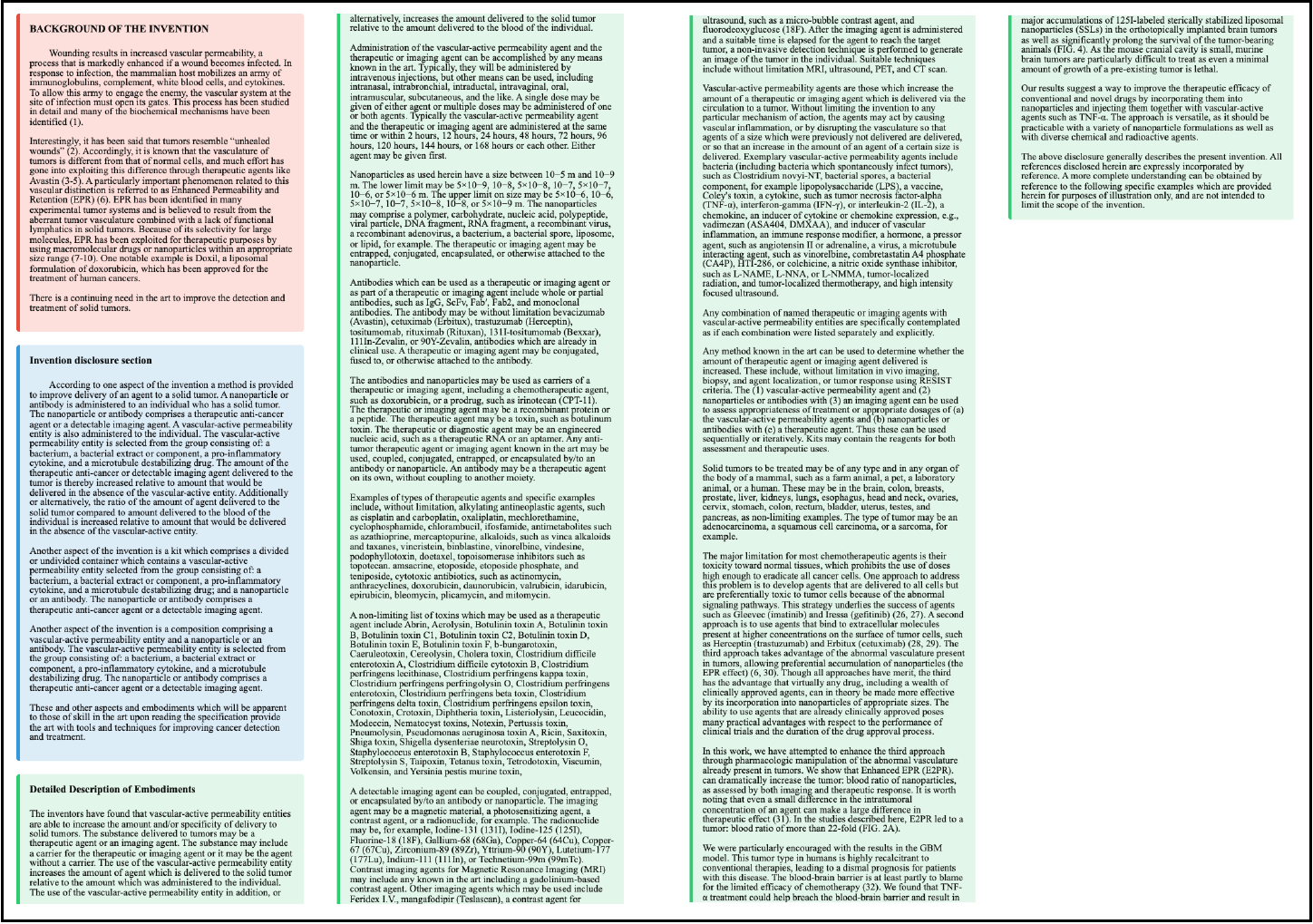}}
\caption{Qualitative case study illustrating the ground-truth reference patent.}
\label{fig:case_study_gt}
\end{center}
\end{figure}

To qualitatively assess the structural fidelity of long-form patent generation, we present a case study comparing a patent generated by our method with its ground-truth counterpart, as shown in Figure~\ref{fig:case_study} and~\ref{fig:case_study_gt}. Both documents are visualized using section-level color coding to facilitate direct comparison of document organization and length distribution.
As observed, the generated patent closely matches the reference patent in overall length and preserves a well-balanced allocation across major sections.

\newpage
\section{Discussion on Human Evaluation and Legal Adequacy}
\label{sec:appendix-discussion}

While this work focuses on algorithmic mechanisms for controllable, evidence-grounded long-form patent drafting in an early-stage drafting setting, we emphasize that the proposed framework is intended as a structured drafting assistant rather than a replacement for professional patent attorneys. Accordingly, aspects critical to real-world patent practice, such as legal adequacy, claim dependency correctness, and antecedent basis consistency, ultimately require expert human judgment.
These criteria extend beyond surface-level textual quality and factual alignment, involving nuanced legal interpretation and jurisdiction-specific drafting conventions that remain difficult to reliably capture with automatic metrics.

We view human evaluation by patent practitioners as an essential direction for future validation. A principled evaluation protocol would involve expert assessment of generated drafts along dimensions including (i) legal adequacy of technical disclosure, (ii) correctness and consistency of claim dependencies, and (iii) validity of antecedent basis usage, potentially complemented by qualitative feedback on drafting clarity and completeness.
We believe that the logic-tree-based intermediate representation introduced in this work provides a favorable foundation for such evaluations by explicitly exposing conceptual dependencies and evidence grounding, thereby facilitating systematic inspection and correction by human experts.